\documentclass{egpubl}
\usepackage{eg2018}

 \electronicVersion 

\ifpdf \usepackage[pdftex]{graphicx} \pdfcompresslevel=9
\else \usepackage[dvips]{graphicx} \fi

\PrintedOrElectronic

\usepackage{t1enc,dfadobe}

\usepackage{egweblnk}
\usepackage{cite}
\usepackage{algorithm}
\usepackage[noend]{algpseudocode} 
\usepackage{algpseudocode}

\usepackage{amsmath,amssymb, tabularx}
\title[ DeepSpline: Data-Driven Reconstruction of Parametric Curves and Surfaces]%
      { DeepSpline: Data-Driven Reconstruction of Parametric Curves and Surfaces}

\author[]
{\parbox{\textwidth}{\centering Jun Gao$^{1,2}$\thanks{jungao@cs.toronto.edu}, Chengcheng Tang$^{4}$, Vignesh Ganapathi-Subramanian$^{4}$, Jiahui Huang$^{3}$, Hao Su$^{5}$, Leonidas J. Guibas$^{4}$
        }
        \\
{\parbox{\textwidth}{\centering 
       $^1$University of Toronto; $^2$Vector Institute;
       }
}
\\
{\parbox{\textwidth}{\centering 
       $^3$Tsinghua University; $^4$ Stanford University; $^5$ UC San Diego
       }
}
}

\begin{document}


\teaser{
\vspace{-15pt}
\includegraphics[width=\textwidth]{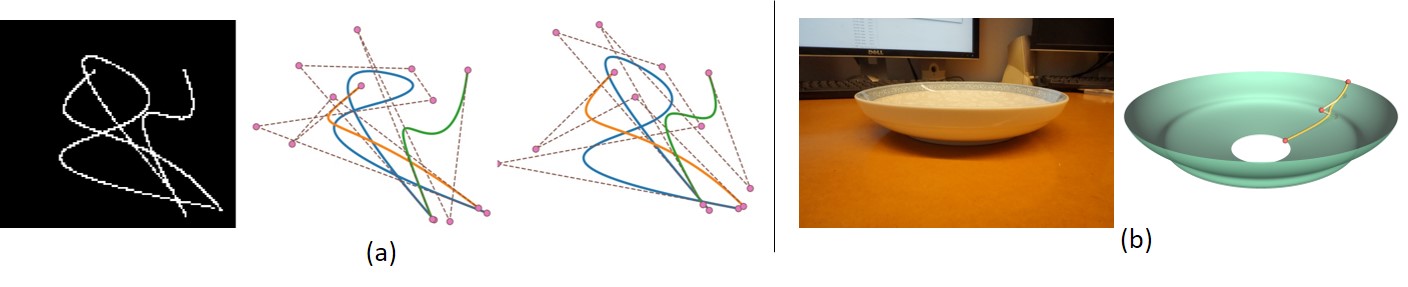}
\centering
\vspace{-20pt}
\caption{ \small 2D and 3D Parametric Reconstruction. (a) Given a binary image as input on the left, we use a Hierarchical RNN to generate its parametric reconstruction as shown in the middle image. This could be used as initialization to obtain a better reconstruction using traditional optimization methods in the right image. (b) A real image is provided as input on the left, and the rotational symmetry of the object in the image is leveraged to reconstruct the surface on the right, after learning the spline curve that models the profile of the object in the image. Please note that even the subtle kink on the dish is captured in our reconstruction.}
\vspace{2pt}
\label{fig:teaser}
}

\maketitle



\begin{abstract}
  Reconstruction of geometry based on different input modes, such as images or point clouds, has been instrumental in the development of computer aided design and computer graphics. Optimal implementations of these applications have traditionally involved the use of spline-based representations at their core.  Most such methods attempt to solve optimization problems that minimize an output-target mismatch. However, these optimization techniques require an initialization that is close enough, as they are local methods by nature. We propose a deep learning architecture that adapts to perform spline fitting tasks accordingly, providing complementary results to the aforementioned traditional methods. We showcase the performance of our approach, by reconstructing spline curves and surfaces based on input images or point clouds.


\end{abstract}  
\section{Introduction}



Three-dimensional data, that is useful for everyday geometric representation and design, is essential in modern industry and research.  
Generating, approximating, processing and storing such data succinctly, and yet faithfully, are vital research problems with a long history of study.
While it would be ideal to use three-dimensional data in its most lossless form, the theoretical limit on representability as well as the source of data collection make the mode of representation crucial.
For different applications, varied modes of 3D data representations are used, such as grids, points clouds, meshes and splines.

In the broad spectrum of 3D data representations, the most abstract forms are those based on splines and parametric primitives.
Their natural representability for arbitrarily smooth surfaces makes them the industry standard for computer-aided design.
Therefore, while representations such as point clouds are more accessible, conversion to parametric representations is often necessary.
Generating clean geometric data with parametric surfaces based on observations often requires approximation and fitting based on a distance measure.
However, the requirement for a measurable distance between target and output and an initialization are common limitations for such a family of fitting algorithms.

On another front, creating geometry that could be induced and inferred from single images has historically been a very attractive research problem.
A strikingly exciting direction is geometric inference from indirect inputs, especially single images.
While such tasks, commonly known as Shape from X, have been studied for over four decades, the recent advances in deep learning have shed a new light on recovering these geometries.
By synthetically generating observations based on ground-truth geometry, recovering geometry based on single images has been illustrated with the representation of voxel grids or point clouds over a wide variety of different shapes.

A natural question that then arises is if there is a way to infer the parametrizable surfaces, directly based on the input information.
A deep learning framework is best suited to perform such inference for multiple reasons.
Fundamentally, deep learning-based frameworks are adaptable to varied input formats, especially when the relationship between the data and geometry is highly nonlinear and non-convex, e.g., images.
Besides, by learning over a myriad of examples, deep networks also do away with the need for manual initialization of the control points, thereby making the inference process less laborious.
Finally, the deep learning frameworks also aid in performing inference when the number of control points and curves are variable in number, which often needs to be determined heuristically in the traditional setting.

In this paper, we attempt to reconstruct spline curves and surfaces using data-driven approaches. 
To tackle challenges with the 2D cases such as multiple splines with intersections, we use a hierarchical  Recurrent Neural Network (RNN) trained with ground truth labels, to predict a variable number of spline curves, each with an undetermined number of control points.
In the 3D case, we reconstruct surfaces of revolution and extrusion without self-intersection through an unsupervised learning approach, that circumvents the requirement for ground truth labels.

To summarize, our contributions are as follows:

\begin{itemize}
\item We define two single-layer RNNs, Curve RNN and Point RNN that can be used to perform curve predictions and control point predictions respectively.
\item We provide a Hierarchical RNN architecture model that performs reconstruction of 2D spline curves of variable number, each of which contains a variable number of control points using nested Curve RNN and Point RNN units, and an algorithm to train it effectively.
\item We provide an unsupervised parametric reconstruction model that performs reconstruction of 3D surfaces of extrusion or revolution. 
\end{itemize} 
\section{Related Work}
\label{related_work}
The problem of recovering faithful yet succinct representations of geometry, has been studied extensively over the past decades with varied forms of outputs and inputs. There are three categories of works that have been relevant to this study and fundamental to multiple applications. We first discuss previous work on spline fitting, in which a direct minimization of a distance between the target and the result is performed. Next, we discuss the creation of shapes from indirect information, i.e., Shape from X, with current advances in deep learning. Finally, we also review the work done in vectorization of rasterized images and discuss the main differences. 

\textbf{Spline Fitting} In computer-aided design and computer graphics, registering or fitting curves or surfaces to targets, e.g., point clouds, are essential for a wide range of tasks such as industrial design following a physical sculpture. One of the most widely used method is the Iterative Closest Point (ICP)~\cite{besl1992method,chen1991object}, which minimizes the distances between two clouds of points based on an initial configuration, iteratively evaluated. Following a similar goal of minimizing a directly measurable and differentiable distance, multiple variants of registration and fitting algorithms are proposed. By devising a metric that is adaptive to curvature, Wang et al proposed a faster and more robust algorithm~\cite{wang2006fitting}, which was further accelerated with quasi-Newton methods~\cite{zheng2012fast}This has further been extended to spline surfaces with constraints such as developability~\cite{tang2016dev}. Despite being able to successfully and efficiently minimize the energies encoding distances, a proper initialization -- a given number of points at selected positions -- is always necessary for such approaches. In contrast, our method does not require such an initialization and could even be used to complement those previous approaches to improve their fits. 

\textbf{Shape from X} There has been a long history of interest in discovering shapes from indirect inputs on images, e.g., from shading~\cite{horn1970shape,horn1989shape} or texture~\cite{kender1979shape}.
Most traditional methods follow a sequential procedure including, e.g., light source estimation, normal estimation, and depth estimation.
More recently, multiple works generate shapes based on images in an end-to-end manner with the help of deep neural networks and synthetically generated data, for shapes represented as volumetric grid~\cite{choy20163d} or point clouds~\cite{fan2016point}.
Despite the variety and complexity of recoverable shapes from single images, a large gap remains between the reconstruction and a clean geometry. 
While multi-scale approaches, e.g., based on octree~\cite{ogn2017}, attempt to enhance geometric details with higher resolution, we attempt to directly recover the  parameters of geometric primitives, especially spline curves or surfaces directly, which provide a higher level geometric abstraction with an arbitrarily high resolution.

\textbf{Image Tracing} Another field of related work is image tracing or vectorization, in which a rasterized image is converted to a vectorized one.
Commercially available tools, e.g., Illustrator, often provide a fine tessellation with an excessive amount of control points to ensure fidelity.
While most vectorization techniques work on the boundaries of the shapes, recent works such as~\cite{favreau2016fidelity} and~\cite{simo2016learning} strive for a simplification of the output on curves in a globally consistent manner.
Besides tracing based on direct differences of colors, Polygon-RNN~\cite{PolyRNN} used an RNN to predict the polygon contour of an object on a semantic level, which could be used for instance segmentation or reducing labelling labor.
In contrast to these methods that perform vectorization, we aim to abstract the simplest types of representation, based on general splines instead of polylines or interpolating cubic Bezier curves, and to create 3D surfaces based on images. 





\section{Overview}
\label{overview_sec}

\begin{figure}[t]
  \centering
  \includegraphics[width=1.0\linewidth]{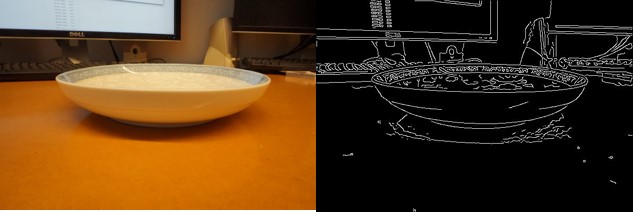}
  \hfill \mbox{}
  \caption{Canny Edge Detector on real image. The image on the left shows a real image with a dish, and the image on the right is the output of the Canny Edge Detector of the image on the former.}
  \label{two_reals}
\end{figure}

\begin{figure}[t]
  \centering
  \includegraphics[width=1.0\linewidth]{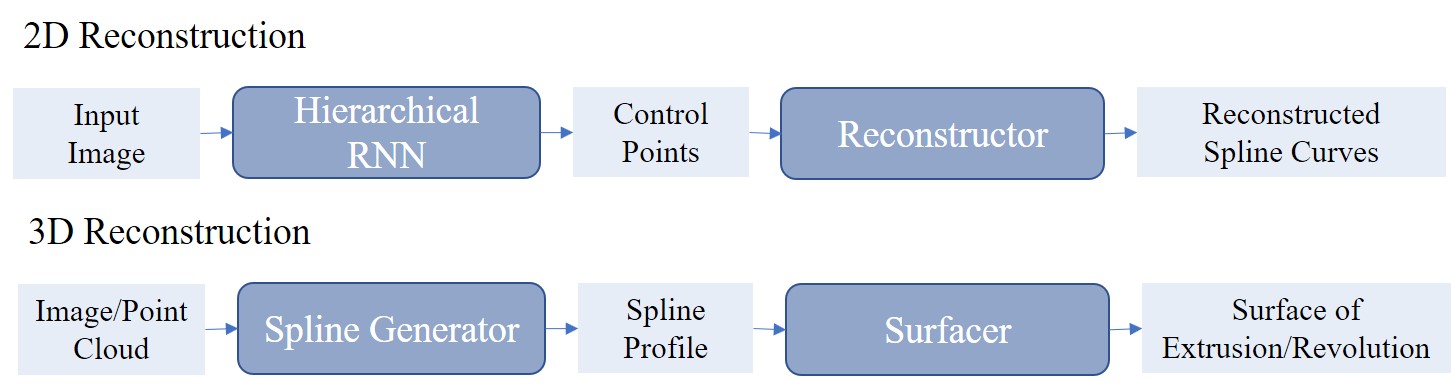}
  \hfill \mbox{}
  \caption{Overview. The image and point cloud representations of curves or 3D shapes are used to generate control points and splines, which can be used for 2D and 3D reconstruction.}
  \label{overview_img}
\end{figure}
\subsection{Motivation}
Traditional methods attempt to extract curved lines or surfaces from images leveraging low-level local image features such as gradients. For example, Canny Edge Detector \cite{Canny:1986:CAE:11274.11275} has been a popular edge detector in computer vision research for decades. Although low-level image features can work well for simple cases when the background is simple and clean, like handwriting on white clean papers, in images with clutter as in Figure~\ref{two_reals}, it is much more challenging to extract the profile of objects. It becomes vital then to develop a more robust method that can better exploit image content. As shown in recent computer vision papers for object recognition ~\cite{NIPS2012_4824}, deep learning methods seem to be able to learn object categories agnostic to the nuisance factors such as lighting, background clutter, pose variation etc. We, therefore, resort to deep learning methods to detect and generate parametric curves for 2D/3D reconstruction. This serves as crucial motivation to solve this problem using learning based techniques.

\subsection{Method}
The spline-based reconstruction techniques presented in this paper are performed on multiple input modalities, specifically images and point clouds, and both 2D and 3D reconstruction are discussed in this paper (Figure~\ref{overview_img}). For 2D reconstruction from images, there are two stages, prediction of spline curves in the image to reconstruct, and prediction of the actual control points that reconstruct the afore-mentioned spline curve. In the paper, we address both these reconstructions as individual problems. We solve the prediction of identifying the spline curves, knowing each of the curves have a fixed number of control points, and the prediction of identifying the control points, given a single spline curve, but not knowing the number of control points that are used to generate this curve. The reconstruction techniques used for these individual problems are then utilized along with a hierarchical deep learning module called the Hierarchical RNN, to solve the more general problem of predicting both the unknown quantities. 

For 3D reconstruction, we perform surface reconstruction in the case of an extruded cross-section or rotational symmetry with two input modes, images or point clouds. These image or point cloud data are processed to learn features, from which spline curves that generate the shape are learned. From the spline curve that is learned, a surface of extrusion or revolution is generated by extruding the curve along the path of extrusion or revolving the curve about the axis of symmetry.

\begin{figure}[htb]
  \centering
  \includegraphics[width=0.95\linewidth]{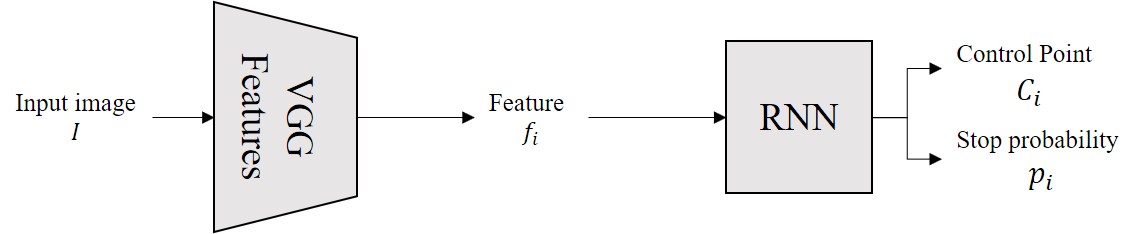}
  \hfill \mbox{}
  \caption{Network architecture for single spline curve fitting with a variable number of control points. VGGNet is used for feature extraction. Image features will be used as input to the RNN. At each iteration $i$, the RNN predicts the position of control point $C_i$ and the stop probability $p_i$ to decide if to continue the iterative process. }
  \label{RNNSpline}
\end{figure}

\section{Supervised 2D Reconstruction}
\label{model}

Reconstructing spline curves in images consists of two modes of variability: the number of spline curves and the number of control points. While solving for both these factors of variability can prove challenging, solving the sub-problems where one of these modes is fixed can provide ample intuition towards solving the harder problems with more variables. In this section, we provide basic models of spline curve fitting solving both these variability issues. First, we propose a model that infers a variable number of control points to fit a single spline curve. Then, we extend this to fitting multiple spline curves with a fixed number of control points. Finally, we propose the Hierarchical RNN to solve for multiple spline curves with a variable number of control points. 

\subsection{Single Spline Curve, Variable Number of Control Points}
\label{basic_network_sec}

A vanilla model is used to tackle the problem of fitting a variable number of control points to a single spline curve. For the spline curve in consideration, the corresponding control points form an ordered sequence, with each element of the sequence being the position of one control point. The prediction of these variable number of control points could, therefore, be viewed as inferring a variable length sequence. A very similar learning technique has previously been used in Machine Translation and Image Captioning. \cite{sutskever2014sequence, bahdanau2014neural, cho2014learning, vinyals2015show}. The use of RNNs for this generative process is a natural choice. 

The input to the pipeline is an image that contains the spline curve. A deep convolutional network is used to extract a feature vector from this image. This feature vector is then forwarded to an RNN module, which predicts the control point sequence. The RNN module performs a dual prediction task. At each iteration, the RNN predicts the position of a new control point and the probability with which this control point is the endpoint. The probability is predicted as a distribution over two states \{CONTINUE = 0, STOP = 1\}.  Specifically, at time step $t$, the model predicts the $t^{\text th}$ control point $C_t$ and the probability $p_t$ for the prediction to stop at this time step. In the ideal scenario, the value of $p_t$ would be binary, with value $0$ for all time steps, and $1$ for the final time step, forcing the prediction to end.

The network architecture is shown in Figure~\ref{RNNSpline}. Here, we use the VGGNet, described in \cite{simonyan2014very},to extract image features, and then perform mean pooling to obtain a vector representation of the whole image. This feature vector is fed into a linear layer to get a more abstract feature vector with 512 dimensions and then supplied into the RNN.  We use a one-layer Gated Recurrent Unit (GRU) \cite{cho2014properties} as the basic block of the RNN, with the dimension of input and hidden layers set to 512. At each time step, the hidden vector $h_t$ of RNN is fed forward into a two-layer fully-connected network to produce the control point's position $C_t$ and the stop probability $p_t$. Specifically, the output of this two-layer fully-connected network has four units, the first two represent $C_t$ and we append a softmax layer into the last two units, which provide a probability distribution over \{CONTINUE = 0, STOP = 1\}.

We use a Mean-Squared Error (MSE) term to optimize on the position of the control points, and a Cross-Entropy term to optimize for the predicted stop probability. The loss function used to train the RNN is 
\begin{align}
\label{loss_variable}
L_1 &= \sum_{i=1}^{N} \Big(\sum_{k=1}^{n_i}(C^i_k - \hat{C}^i_k)^2 \\
    &+ \lambda \big(p^i_k\log(\hat{p}^i_k) + (1-p^i_k)\log(1-\hat{p}^i_k)\big) \Big)
\end{align}
where $N$ is the size of the training dataset, $n_i$ is the number of control points of $i^{\text{th}}$ spline curve in the training dataset, $\hat{C}$ is the predicted position of a control point, while $C$ is the corresponding ground truth, $\hat{p}$ is the predicted stop probability, $p$ is the ground truth stop probability ($1$ when $k = n_i$, $0$ otherwise) and $\lambda$ is the optimization hyperparameter.

\begin{figure}[htb]
  \centering
  \includegraphics[width=.95\linewidth]{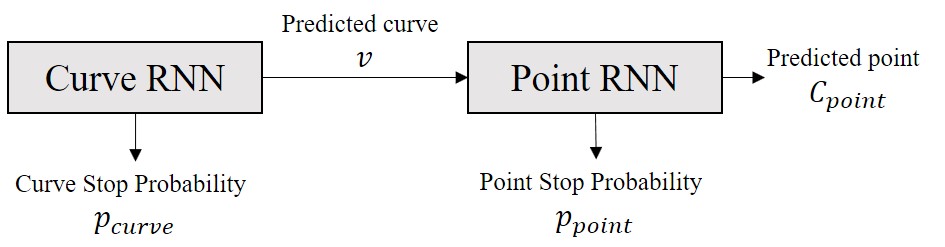}
  %
  %
  \caption{\label{hierarchical_rnn_model}
            Hierarchical RNN architecture. The Curve RNN acts as an outer loop to determine when all curves in the image have been generated. For each iteration of the Curve RNN, the Point RNN runs the prediction based on the Single Curve, Variable Control Points process as described in Section \ref{basic_network_sec} with input obtained from the Curve RNN to perform reconstruction of this curve. The Curve RNN predicts the stop probability to signal end of all curves, $p_{\text{curve}}$ and the predicted curve $v$, while the Point RNN predicts the stop probability to signal end of predicted points, $p_{\text{point}}$ and the predicted point at that particular iteration $C_{\text{point}}$.}
\end{figure}

\begin{figure*}[tbp]
  \centering
 \vspace{-20pt}
  \mbox{} \hfill
  \includegraphics[width=\textwidth]{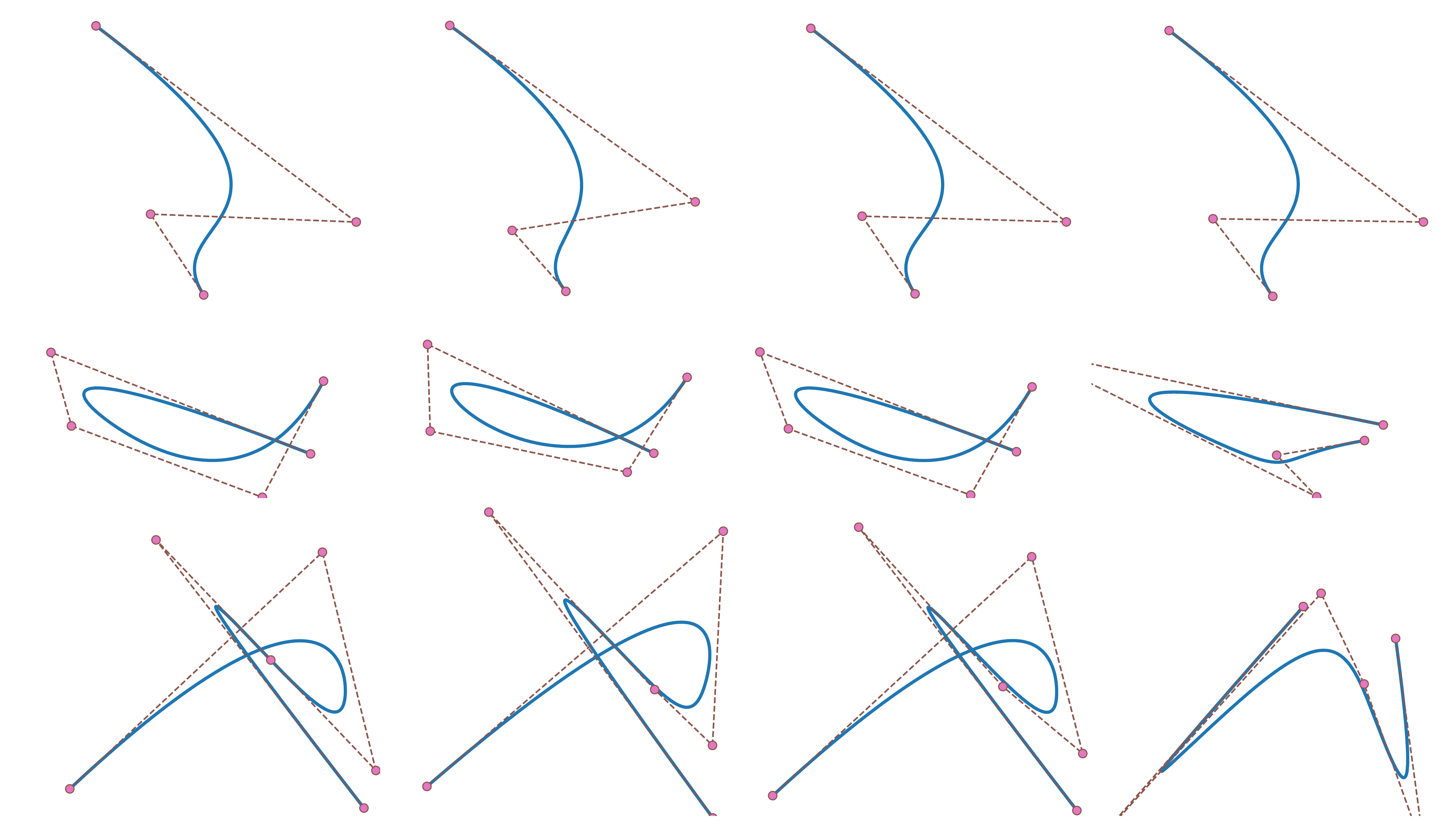} 
  \hfill \mbox{}
  \caption{Randomly picked results of variable number of control points. The first column shows the target curve, the second column shows the prediction from the RNN, the third column shows the reconstruction obtained by using the RNN prediction as initialization for traditional methods, and the last column shows the reconstruction obtained by using a random initialization.}
  \label{results_variable}
\end{figure*}

\begin{figure*}[tbp]
  \centering
\vspace{-20pt}
  \mbox{} \hfill
  \includegraphics[width=\textwidth]{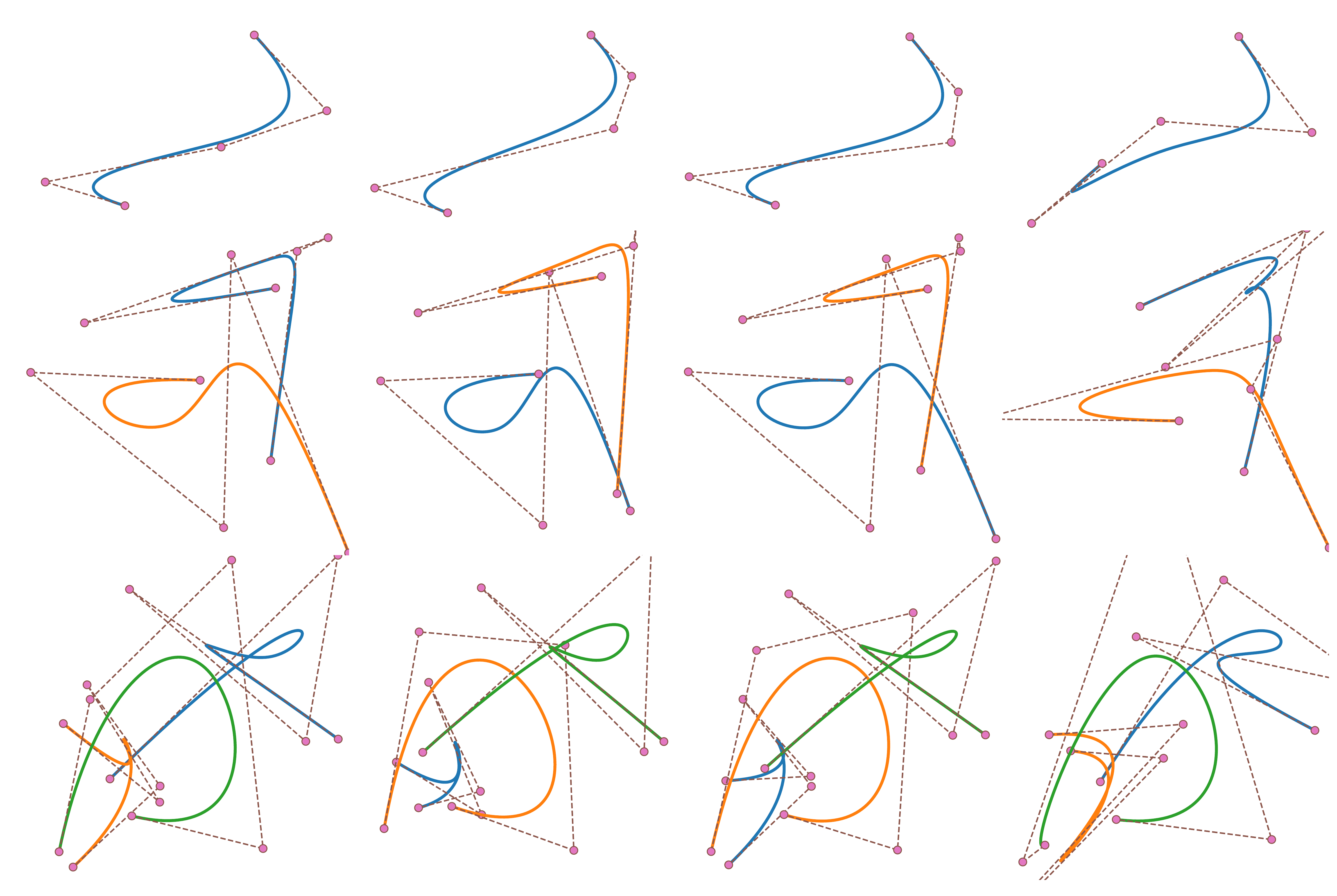} 
  \hfill \mbox{}
  \caption{Randomly picked results of multiple number of spline curves with fixed number of control points. The first column shows the target curves, the second column shows the prediction from the RNN, the third column shows the reconstruction obtained by using the RNN prediction as initialization for traditional methods, and the last column shows the reconstruction obtained by using a random initialization.}
  \label{results_multiple}
\end{figure*}

\subsection{Multiple Spline Curves, Fixed Number of Control Points}
\label{multiple_curves_sec}

Fitting multiple spline curves with a fixed number of control points is solved with a minor modification to the model in Sec.\ref{basic_network_sec}. Here, instead of predicting if a certain point is the final control point, the entire spline curve along with all control points are computed at each iteration of the RNN. The RNN also predicts the probability to determine if the most recent curve is the final curve. 

However, there are two new challenges that arise in this solution. The RNN predicts an ordered sequence of curves, while the target of multiple spline curves is an unordered bag of spline curves. Therefore, a correspondence needs to be established between the target curves and the curves that are predicted by the network. This is achieved by modeling this problem as a matching problem in a bipartite graph, with the two sets being the set of target curves and the set of predicted curves. The weight of each edge between the curves in the two sets would be the distance between the two curves. The Hungarian Algorithm is implemented to obtain a matching of minimal cost. This is similar in spirit to the matching problem solved in \cite{eccv_15_ris}.

The second challenge is to ensure that when a certain spline curve is being processed, influences from regions of other spline curves are minimized. Since there are multiple curves, occupying different regions of the image, it becomes necessary to nullify these influences. This is handled by adapting the attention mechanism \cite{sukhbaatar2015end-to-end, kumar2016ask} of the network. At each time step $t$, the image features are scaled by an attention map showing weights of different regions in the image before passing into the RNN. This ensures that the attention of the network is localized to the region in which the current spline dominates. The idea of using a localization network to perform this task of drawing attention to certain regions over others is used with considerable success in\cite{icml2015_xuc15}. Our methods are demonstrated in more detail in Sections \ref{hungarian_algo} and \ref{attention_module}.

\subsubsection{Loss Function}
\label{hungarian_algo}

The training dataset is composed of labeled images. Image $i$ as $\textbf{I}^i\in \mathcal{R}^{128x128x3}$, its annotation $S^i = \{S_1^i, S_2^i, ... , S_{n_i}^i\}$ is a set of spline curves, $n_i$ is the number of spline curves in this image, $S_j^i$ is a sequence of control point positions that construct spline curve $j$ inside the image $\textbf{I}^i$. In all our experiments, we use $n_i \in \{1, 2, 3\}$.

The network predicts both a sequence of spline curves $\hat{S}^i = \{\hat{S}_1^i,\hat{S}_2^i, ... ,\hat{S}_{\hat{n}_i}^i\}$, and stop probabilities $\hat{P}^i = \{\hat{p}_1^i, \hat{p}_2^i, ... , \hat{p}_{\hat{n}_i}^i\}$. We train the RNN by running $n_i$ iterations for each training instance $S^i$. On inference, the number of iterations is determined by the predicted stop probability $\hat{p}_j^i$. The recurrence stops when $\hat{p}_j^i > 0.5$.

The loss on the probability sequence is defined as earlier. 

\begin{equation}
L_p^i = \sum_{j = 1}^{n_i} p_j^i \log \hat{p}_j^i + (1-p_j^i)\log(1-\hat{p}_j^i)
\end{equation}
\begin{equation}
p_j^i = 
\begin{cases}
0& j<n_i\\
1& j=n_i
\end{cases}
\end{equation}

As described earlier, the loss term that measures the distance between the target and predicted spline curves also needs alignment. Since $\hat{S}^i$ is an ordered sequence and the ground truth $S^i$ is an unordered set, the order of processing the spline curves in an image is not easily determinable. Random allocation of processing rank to $S^i$ will result in a problem of ambiguity and the network convergence is not guaranteed. The bipartite graph model is used here, as described earlier, to model the correspondence, while computing the reconstruction loss. This reconstruction loss is computed as follows: 

\begin{equation}
\label{loss_min_hungarian}
L_c^i = \min_{\delta \in \mathbf{K}^i} f_{Match}(S^i, \hat{S}^i, \delta)
\end{equation}
where
\begin{equation}
f_{Match}(S^i, \hat{S}^i, \delta) = \sum_{j = 1}^{n_i}\big(\sum_{\hat{j} = 1}^{n_i} f_{ED}(S_j^i, \hat{S}_{\hat{j}}^i)\delta_{j,\hat{j}}\big),
\end{equation}

\begin{align}
\label{matching}
\mathbf{K}^i = \Big\{\delta \in \{0,1\}^{n_i, n_i}: &\sum_{j=1}^{n_i}\delta_{j,\hat{j}} = 1, \forall \hat{j} \in \{1,...,n_i\}, \\
&\sum_{\hat{j}=1}^{n_i}\delta_{j,\hat{j}} = 1, \forall j \in \{1,...,n_i\}\Big\}.
\end{align}

$f_{ED}(S_j^i, \hat{S}_{\hat{j}}^i) = \sum_{k=1}^m (C_{S_j}^k - C_{\hat{S}_{\hat{j}}}^k)^2$ is the Euclidean distance between the position of control points at $S_j^i$ and $\hat{S}_{\hat{j}}^i$, and $m$ is the number of control points. we use $m = 5$ in our experiments. 

$\delta$ defines a matching between ground truth $S^i$ and prediction result $\hat{S}^i$, Equation \ref{matching} guarantees a bijection between the ground truth and prediction results. The optimal matching $\delta$ in Equation \ref{loss_min_hungarian} is efficiently computed by the Hungarian algorithm. Therefore, the overall loss function is defined as:
\begin{align}
L_2 &= \sum_{i=1}^{N}\Big( \min_{\delta \in \mathbf{K}^i} f_{Match}(S^i, \hat{S}^i, \delta)\\
    &+ \lambda\sum_{j = 1}^{n_i} \big(p_j^i \log \hat{p}_j^i + (1-p_j^i)\log(1-\hat{p}_j^i)\big)\Big)
\end{align}

\begin{figure*}[tbp]
  \centering
 \vspace{-20pt}
  \mbox{} \hfill
  \includegraphics[width=\textwidth]{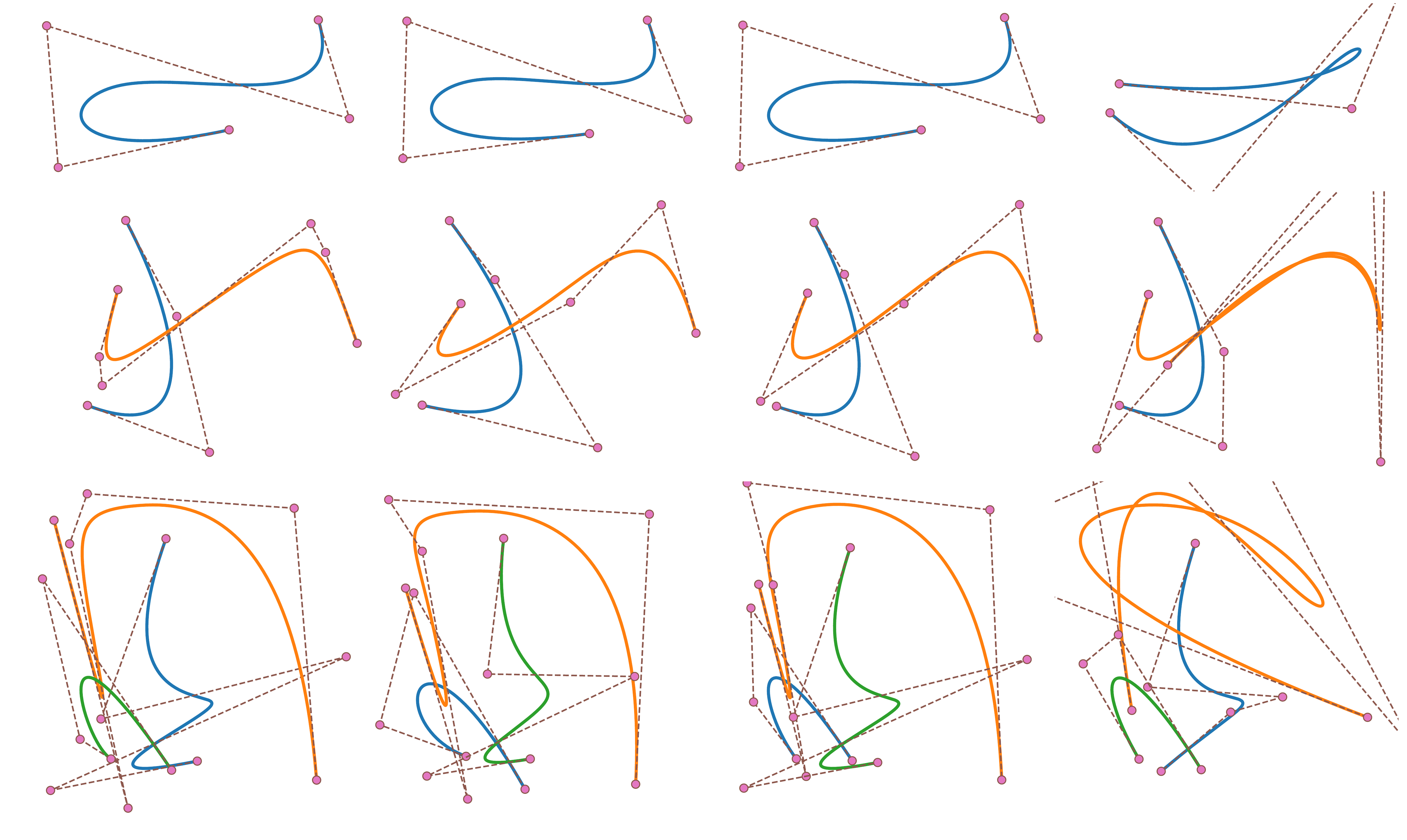}  
  \hfill \mbox{}
  \caption{Randomly picked results of multiple spline curves with variable number of control points. The first column shows the target curves, the second column shows the prediction from the Hierarchical RNN, the third column shows the reconstruction obtained by using the Hierarchical RNN prediction as initialization for traditional methods, and the last column shows the reconstruction obtained by using a random initialization.}
  \label{results_multiple_variable}
\end{figure*}

\subsubsection{Attention Network}
\label{attention_module}
At each iteration of the RNN, a smaller network $f_{\textbf{att}}$ is used to predict a soft attention mask. This mask provides localized weights to different regions of the image features.

The output of the feature extractor is a 3D tensor $\mathcal{M}$, of size $(l,x,y)$, where $l$ is the number of channels in the last layer of the network and $(x,y)$ is the downsampled size of input image. Each column, a vector that has $l$ elements, relates to a certain region in the input image. Let $d = xy$ and the image feature be $\alpha$.
\begin{equation}
\label{img_feature}
    \alpha = \{ \textbf{a}_1, \textbf{a}_2, ... , \textbf{a}_d\}, \textbf{a}_i \in \mathbb{R}^l
\end{equation}
At each time step $t$ of the RNN, we denote the input vector as $\textbf{z}_t$ and hidden vector as $\textbf{h}_t$ 
\begin{equation}
    \textbf{h}_t = RNN(\textbf{z}_t, \textbf{h}_{t-1})
\end{equation}
The attention network $f_{\textbf{att}}$ is used to get the vector $\textbf{z}_t$, the hidden vector $\textbf{h}_t$ is used to predict the control points and stop probability. 
\begin{align}
\label{attention_function}
    e_{ti} &= f_{\textbf{att}}(\textbf{a}_i, \textbf{h}_{t-1}) \\
    \textbf{z}_t &= \sum_{i=1}^d e_{ti} \textbf{a}_i
\end{align}
$f_{\textbf{att}}$ is a two-layer fully-connected network with ReLU activation. The inputs are $\textbf{a}_i$, which corresponds to a certain part of the image, and $\textbf{h}_{t-1}$, which contains the information of the curve that needs to be predicted at current step. The output is only a scalar $e_{ti}$ which corresponds to the information about how important region $i$ is.


\subsection{Multiple Spline Curves, Variable Number of Control Point}
\label{hierarchical_rnn_sec}

Extending the previous model to the generation process of multiple spline curves with variable numbers of control points would consist of two nested loops. First, we loop over the curves and determine the number of curves, thereby generating one curve at each iteration of the generative process. The inner loop uses the information from the first loop that is needed to generate one curve, looping over the control points and determine the number of control points to generate the curve. This is implemented through a hierarchical RNN structure that is explained below.

\subsubsection{Hierarchical RNN}
\label{hierarchical_rnn}

We propose a Hierarchical RNN structure to model this generation procedure. This architecture has been previously used to describe visual content by attempting to understand and caption details in local regions of an image \cite{krause2016paragraphs}. We leverage the multiple recurrent units of a network to repeatedly extract local regions of an image containing information, and then to process this information. As before, an image feature extractor and an attention subnetwork are used to process the input image before feeding it to the RNN. This model constitutes two RNNs combined hierarchically, one for looping over the curves (Curve RNN) and the other for looping over the control points (Point RNN). At each time step of the Curve RNN, it predicts the stop probability of the curve generation and also generates a vector representation of the current spline curve and forward it into the Point RNN. The Point RNN uses this vector representation of a spline curve and then decodes it into the position of control points. Our model is shown in Figure~\ref{hierarchical_rnn_model}.

\subsubsection{Curve RNN}

The Curve RNN is a single-layer GRU with hidden size $H = 512$, the initial hidden vector $\textbf{h}_0$ is predicted by a two-layer fully-connected network with the average image feature as input. At each time step, the Curve RNN receives the image feature vector $\textbf{z}_t$, after passing through the attention subnetwork, as input. The hidden vector $\textbf{h}_t$ is then fed into a two-layer fully-connected network to obtain curve stop probability $p_t$ and a vector representation  $\textbf{v}_t$ using which the model predicts the curve at the current step, which is also the input to the Point RNN. This is different from what is done in Sections \ref{basic_network_sec} and \ref{multiple_curves_sec}, where the hidden vector is directly fed into the two-layer fully-connected network to get positions of control points (or control point sequence).

\subsubsection{Point RNN}
The Point RNN is also a single-layer GRU with hidden size $H = 512$, which, given a vector representation $\textbf{v}_t$ from Curve RNN, is used to generate the sequence of positions of control points. We follow the network configuration in Section \ref{basic_network_sec}. At each time step, the Point RNN predicts one control point combined with a stop probability that represents whether this control point is the end point.

\subsubsection{Training and Inference}
The two RNNs are trained accordingly to predict the spline curves, and their stop probabilities as suggested in previous sections. The pseudocode to train this Hierarchical RNN model is provided below in Algorithm \ref{algo_hierarchical_rnn}.

\begin{algorithm}  
\caption{Pseudocode for Training Hierarchical RNN}
  \begin{algorithmic}[1] 
  
    \For{$\text{epoch} = 0 \to T$} \Comment{Loop over epoch}
      \For{$i = 0 \to N$} \Comment{Loop over all training data}
        \State $\alpha = CNN(\textbf{I}^i)$ \Comment{Get the image feature}
        \State $\textbf{h}_{curve} = Hidden\_pred_{curve}(\alpha)$ \Comment{Initial hidden vector}
         
        \For{$j = 0 \to L$}\Comment{Predict the attention mask}
          \State $e_{j} = f_{\textbf{att}}(\textbf{a}_j, \textbf{h}_{curve})$
        \EndFor
        \State $\textbf{z} = \sum_{j=1}^D e_{j} * \textbf{a}_j$ \Comment{Get the input vector for RNN}
        \For{$curve = 0 \to n_i$} \Comment{Loop over spline curves}
          \State $\textbf{h}_{curve} = Curve\_RNN(\textbf{z}, \textbf{h}_{curve})$
          \State $\textbf{v} = Curve\_to\_Point(\textbf{h}_{curve})$
          \State $p\_curve= Stop\_Prob(\textbf{h}_{curve})$ \Comment{stop probability}
          
          \For{$j = 0 \to D$} \Comment{Predict the attention mask}
            \State $e_{j} = f_{\textbf{att}}(\textbf{a}_j, \textbf{h}_{curve})$
          \EndFor
          \State $\textbf{z} = \sum_{j=1}^D e_{j} * \textbf{a}_j$
          \State $\textbf{h}_{point} = Hidden\_pred_{point}(\textbf{v})$ \Comment{Initial hidden}
          \For{$point = 0 \to n\_point$} \Comment{Loop over Point}
            \State $\textbf{h}_{point} = Point\_RNN(\textbf{v}, \textbf{h}_{point})$
            \State $C_{point}, p_{point} = Point\_Pred(\textbf{h}_{point})$
          \EndFor
        \EndFor
      \EndFor
    \EndFor
  \end{algorithmic}
  \label{algo_hierarchical_rnn}  
\end{algorithm}  

The loss function has three terms: the mean-squared error of predicted positions, the cross entropy loss of curve stop probabilities and of point stop probabilities:
\begin{align}
L_3 &= \sum_{i=1}^{N}\Big( \min_{\delta \in \mathbf{K}^i} f_{Match}(S^i, \hat{S}^i, \delta)\\
    &+ \lambda_1\sum_{j = 1}^{n_i} \big(p_j^i \log \hat{p}_j^i + (1-p_j^i)\log(1-\hat{p}_j^i)\big) \\
    &+ \lambda_2\sum_{j = 1}^{n_i}\sum_{k=1}^{n_{i,j}} \big( p_{j,k}^i \log \hat{p}_{j,k}^i + (1-p_{j,k}^i)\log(1-\hat{p}_{j,k}^i)\big)\Big) 
\end{align}
$\lambda_1$ and $\lambda_2$ are optimization hyperparameters. $n_i$ is the number of spline curves in the $i^{\text{th}}$ image, $n_{i,j}$ is the number of control points in the $j^{\text{th}}$ spline curve of image $i$. $p$ is the target probability and $\hat{p}$ is the predicted probability.

\section{Unsupervised Parametric 3D Reconstruction}
\label{unsupervised_sec}

While the methods discussed in Section \ref{model} could be employed to perform control point prediction when corresponding ground truth exists, solving for the parameters in the absence of ground truth is significantly harder. 

Two natural obstacles are to be overcome in this setting. The first is to devise a loss function different from Mean-Squared Error that can be used to optimize the neural network. This is because one needs a ground truth to make a point-wise comparison and compute the error. This ground truth is missing in this case, and a technique to compare against a target point cloud, must be devised. We assume we have the target point cloud, which is natural and necessary in traditional methods. The second obstacle is to make the network aware of its purpose, that is to predict spline curves (or surfaces), as opposed to some other arbitrary primitives. These obstacles are entangled two-way, with the loss function design needing to take into consideration properties of spline curves (or surfaces), which might provide feedback to the neural network to determine the parameters.

Both traditional optimization methods (ICP ~\cite{besl1992method,chen1991object}) and ideas in \cite{pmlr-v70-tompson17a}, which utilizes neural network for fluid simulation, provide as inspirations to solving this problem. Suppose we are provided an oracle which could predict the parameters perfectly, then the point cloud representation of the predicted curves (or surfaces) could be generated. The Chamfer distance measurement between the point clouds of predicted curves (or surfaces) and the target point clouds could be used as the loss function to train the network.

Though we use the same loss function as in the case of traditional methods, we use a learning technique to predict the control points, as opposed to optimizing them directly, since this helps us leverage the entire dataset, as opposed to considering just one sample while optimizing. The prediction learned by the network could also used as the initialization to the traditional optimization methods, circumventing the manual initialization while reducing the number optimization iterations. A learning based approach also equips us to deal with multiple input formats, while traditional methods only use point cloud input.

\subsection{Reconstruction of Images or Point Clouds}
\label{real_recons_sec}
In this section, we provide a technique to reconstruct surfaces of extrusion or revolution from images or point clouds. For generating a surface of extrusion, we extrude a spline curve at a random height. As for a surface of revolution, this is generated by revolving the spline curve around the axis by 360 degrees. We only consider spline curves without self-intersections and with monotonically decreasing $y-$coordinate in the control point sequence. This assumption is made since generally self-intersected surfaces are not ubiquitous. Given the input data, features are extracted from it. If this is a 2D image, the VGG network \cite{simonyan2014very} is used to extract the image features as in Section~\ref{basic_network_sec}. If the input is a point cloud, the PointNet \cite{qi2016pointnet} is used to extract features. The extracted features $\textbf{v}$ are forwarded into a two-layer fully-connected network, which predicts the position of control points $\textbf{C}$ (if surfaces of extrusion are considered, $\textbf{C}$ also contains the extruded height). Since $\textbf{C}$ contains the control points, the predicted curve can be reconstructed from a linear combination of $\textbf{C}$, which is dependent on the parameterization of the spline curve $\textbf{t}$. This is represented as follows.

\begin{equation}
\textbf{C} = NN(v)
\end{equation}
\begin{equation}
P_{pred} = \textbf{f}(\textbf{t})(\textbf{C})
\end{equation}
where $NN$ denotes the two-layer fully-connected network, and $\textbf{f}$ is the basis function dependent on the spline generator $\textbf{t}$.

We use the Chamfer distance to measure the distance between the predicted point cloud and target point cloud.
\begin{equation}
d_{CD}(P_1, P_2) = \sum_{x\in P_1}\min_{y\in P_2} \|x-y\|_2^2 + \sum_{y\in P_2}\min_{x\in P_1} \|x-y\|_2^2
\end{equation}

Since $d_{CD}$, is differentiable, it is possible to train the network end-to-end. The function $\textbf{f}(\textbf{t})(.)$ represents control point weights to generate the surface. This is implemented as a linear layer with fixed weights in the network. This architecture is generalizable, since predicting other kinds of surfaces (like surfaces of sweeping or NURBS), would require only a change of this individual layer, with the rest of the model remaining the same.

\section{Experiments}

Training deep neural networks usually requires a copius amount of data. However, since there isn't enough real data with ground truth spline curve labeling, we use randomly synthesised data to perform our training and testing experiments. We synthesize a dataset of size 500,000, with a 70-30 train-to-test split. For each instance of the dataset, multiple spline curves and control points for each of the spline curves are generated. The training dataset consists of self-intersecting or looping spline curves, but do not contain any closed spline curves. This could be done by adding a circular loss term to the loss function in Equation \ref{loss_variable}. But for the purpose of this paper, we shall not venture into closed curves. According to the problem we attempt to solve, we generate the dataset and perform the training. We also run the trained model on real images to check the generalizability of our model. 

For 2D reconstruction, we set the image size to 128$\times$128 and randomly generate the number of spline curves and the number and the position of control points for each spline curve in 2D image plane. The number of control points are varied between 4 to 6 for the problem with spline curve reconstruction with variable control points. The number of spline curve are varied between 1 to 3 for the problem of reconstructing multiple spline curve with the number of control points fixed at 5. These two variations are combined to train and test for multiple spline curve reconstruction with variable number of control points. For 3D reconstrunction, the number of control points are fixed to 5, and then the spline curve is revolved or extruded to gain a 3D surfaces. If the input is an image, we render the 3D surface with fixed camera angle and lighting condition. If the input is point cloud, we randomly sample from the 3D surface.

We use the VGGNet as image feature extractor for all the three models in Sec. \ref{model}. We initialize our network using pretrained weights from ImageNet \cite{deng2009imagenet}. We then train the network with Adam optimizer through back-propagation. We set both the learning rate and weight decay to $10^{-4}$. Determining the number of points (or curves) is easier than predicting the position of the control points. Therefore, the hyperparameter $\lambda$ is set to 0.1 to lay more emphasis on position regression. Our models are implemented in PyTorch and trained for 100$\times$10000 iterations with batch size set to 32. It took approximately one to two days on GeForce GTX 1080 GPU to converge. Testing on one image needs only 12ms.

Measuring the performance of our model is not a straightforward task. Due the ambiguity that could potentially be caused by two shape-similar curves having totally different control point sequences, or even different number of control points, one single measurement would not be able to cover all the characteristics of spline curves. We use three measurements here that complement each other:  Mean-Squared Error between the position of predicted control points and target control points, Classification accuracy of the number of control points and the number of spline curves and Chamfer distance between point clouds of the predicted spline curves and the target curves.

\subsection{Baseline Method}
Traditional methods minimize a distance metric measured directly between a pair of geometric entities: the reconstructed surface and a target often provided as a point cloud. In each iteration, the target points are projected onto the predicted surface (and vice versa) to obtain point-to-foot-point matches. As this distance can be expressed through the parameters of the prediction, minimizing the distance updates the parameters. Throughout the optimization, at each iteration, the distance and the foot-points are reevaluated. Faster algorithms based on normal-to-normal distances and point-to-tangent-plane distances have been used to accelerated the process, with an essentially similar idea of minimizing a local distance metric~\cite{besl1992method,chen1991object,wang2006fitting,tang2016dev}. For all our experiments, we minimize the point-to-point distance as the traditional method to compare against.

\subsection{Results}
\label{results_sec}
We present results pertaining to both 2D spline curve reconstruction and 3D surface of extrusion or revolution reconstruction. As mentioned in Section~\ref{model} and~\ref{unsupervised_sec}, the method that is employed to perform the 2D spline reconstruction is supervised, while the one used to reconstruct the surface of extrusions or revolutions is unsupervised. 

\subsubsection{Supervised 2D Reconstruction}
We have proposed three supervised models in Sections~\ref{basic_network_sec},~\ref{multiple_curves_sec} and ~\ref{hierarchical_rnn_sec} aiming to reconstruct single spline curve with variable control points, variable number of spline curves with a fixed number of control points and variable number of spline curves with variable control points respectively. In this section, we showcase the reconstruction performance of these individual models, as well as in comparison to traditional energy-based models.

\textbf{Single Spline Curve, Variable Number of Control Points.} \\
We attempt to show that traditional energy-based models that optimize to reconstruct a spline curve with fixed number of control points as input, show significantly improved performance when initialized with control points that are more systematically learned by our method, as opposed to a random initialization that is classically performed. To this effect, we randomly initialize the position of control points and use energy-based methods to perform the optimization. Since the traditional method uses a fixed number of input control points, we perform this optimization for three different cases of initial control points ($4,5,6$) and select the best optimized case as the final result. We observe that the performance of the learned initialization easily outperforms the random initialization. This is expected, since the systematic learning of the control points leverages the information contained in the training datasets about the relative position of the control points to the curves, and this can provide us with a very good starting point to perform the optimization from, This is observed in Figure~\ref{results_variable}. It is observed that the prediction by our RNN is close enough to the original curve, and therefore serves as an excellent initialization for control point optimization. After optimization based on this initialization, the obtained reconstruction almost exactly mirrors the input curves. 

\textbf{Variable Spline Curves, Fixed Number of Control Points.} \\
The method discussed in Section~\ref{multiple_curves_sec} is used to perform predictions of number of spline curves and the control points for each of the spline curves. In Figure~\ref{results_multiple}, we again perform the comparisons between the target and the predictions of the RNN, the optimization-based reconstruction dependent on initializations, both random and based on RNN output. We again observe that the prediction of the RNN, when used as an initialization, comfortably outperforms the random initialization.

\textbf{Variable Spline Curves, Variable Number of Control Points}\\
The method discussed in Section~\ref{hierarchical_rnn_sec} is used to perform predictions of a variable number of spline curves, each containing a variable number of control points. We perform the comparisons between the target and the random and RNN-based initialization and reconstruction schemes again in Figure~\ref{results_multiple_variable}. It is also to be noted that in  Figure~\ref{results_multiple_variable}, as in the case of the orange curve in the second row, in spite of prediction of number of control points being wrong and predicted positions also being wrong, the predicted and target curves look similiar, which provide us with the intuition that the RNN architecture learns the curves though it does not necessarily learn it in the specific manner we intend it to. 
\begin{figure}[!htb]
  \centering
  \mbox{} \hfill
  \begin{minipage}{\linewidth}
    \includegraphics[width=0.32\linewidth]{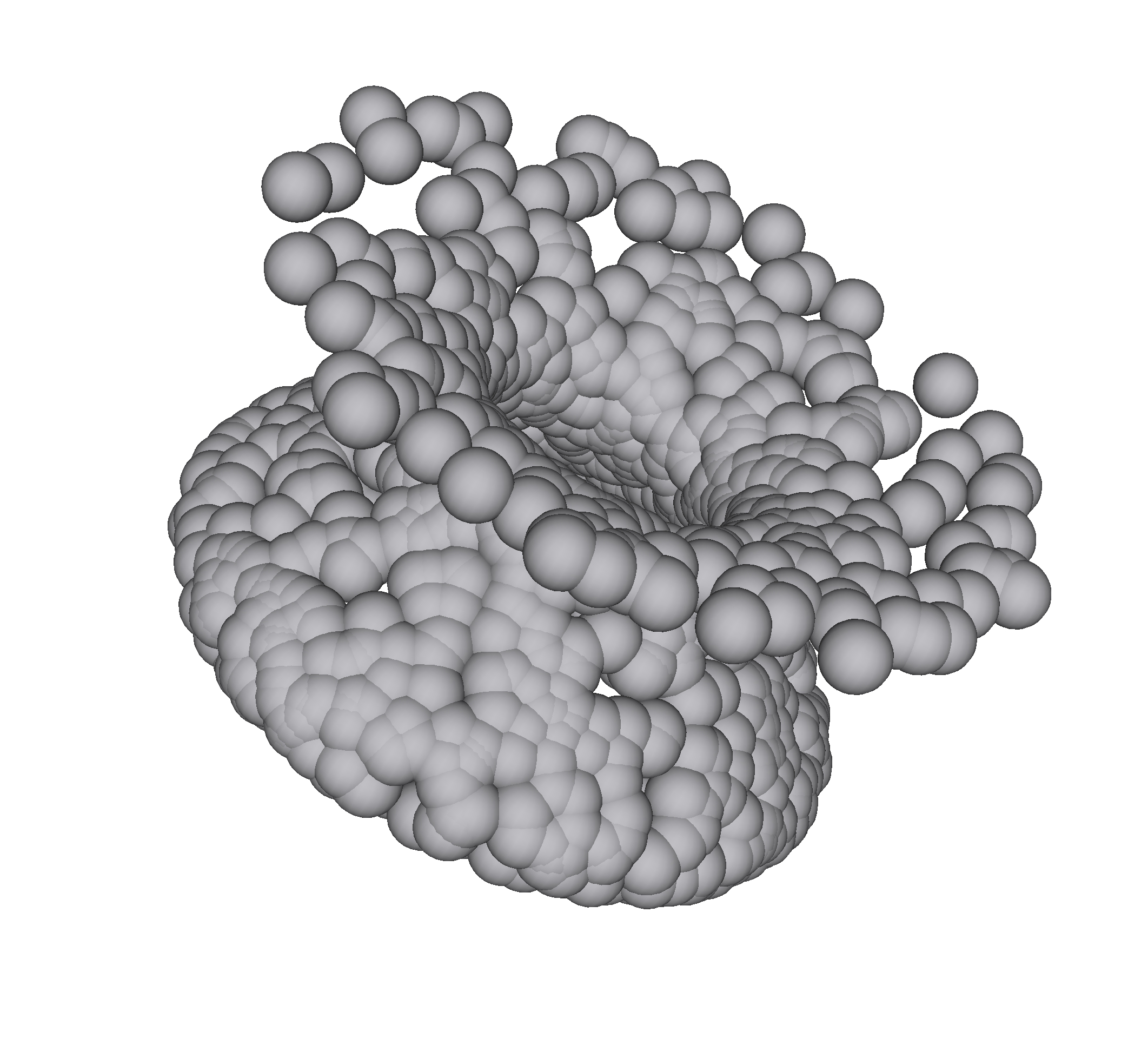}
    \includegraphics[width=0.32\linewidth]{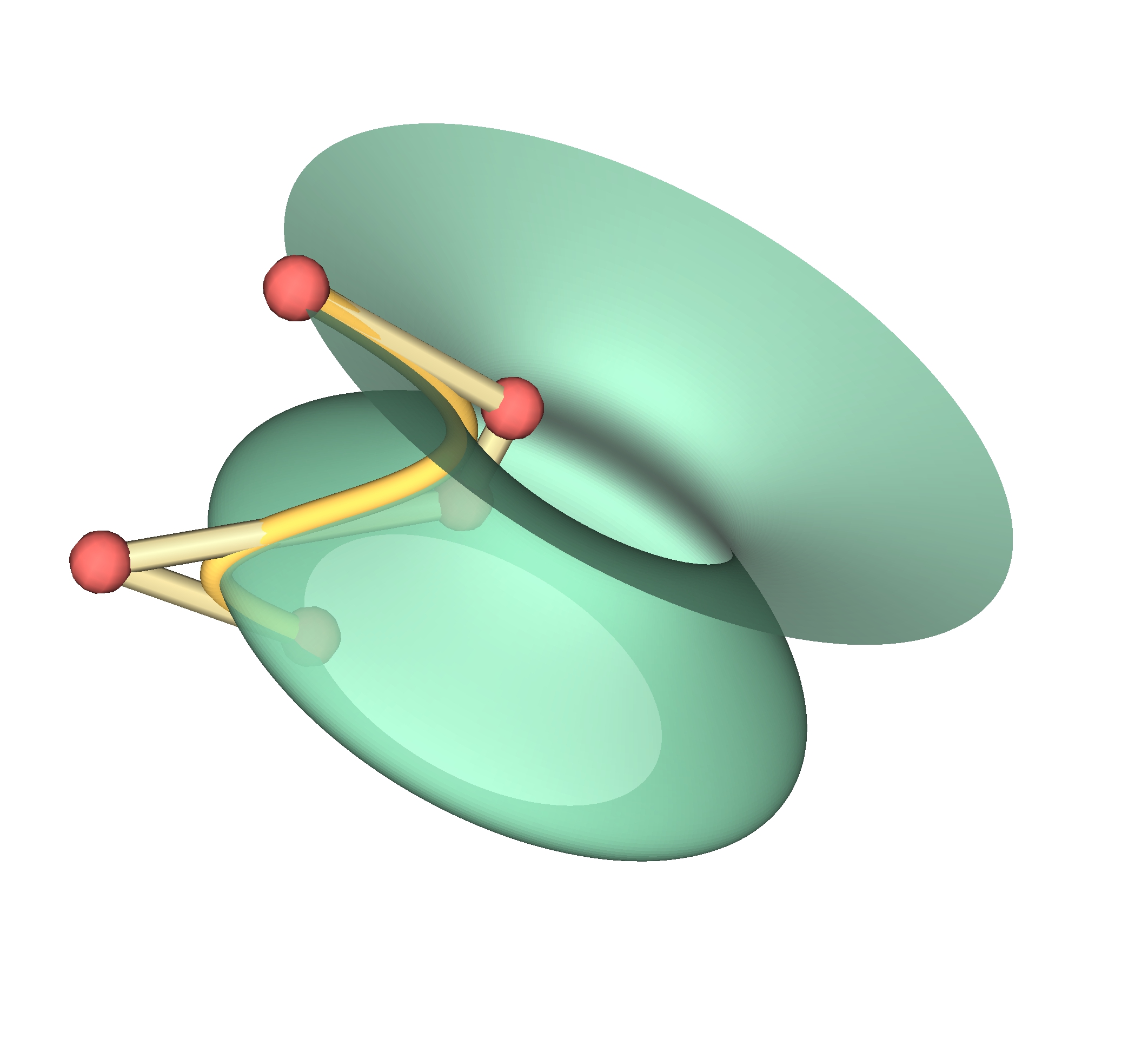}
    \includegraphics[width=0.32\linewidth]{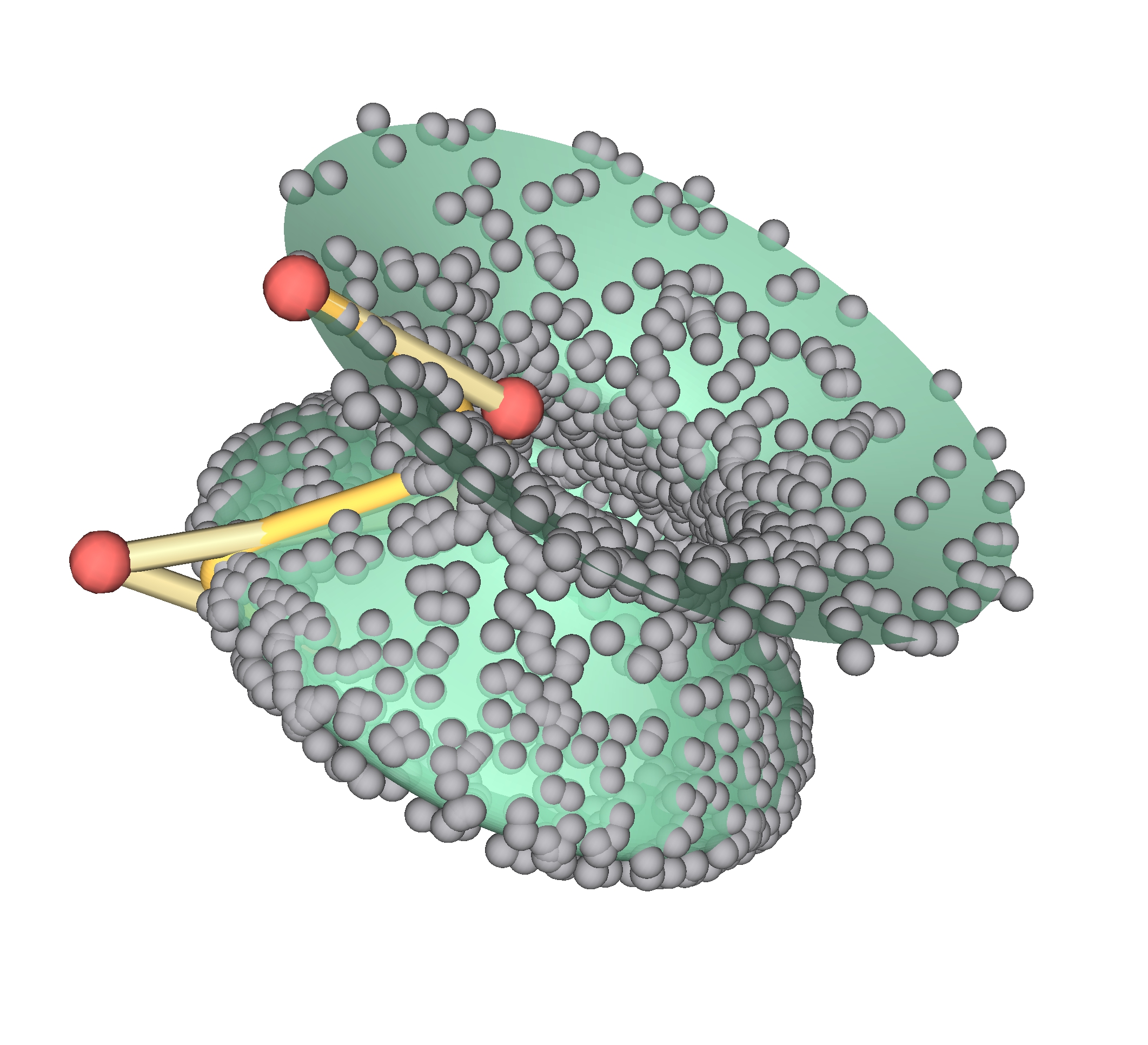}
  \end{minipage}
  \begin{minipage}{\linewidth}
    \includegraphics[width=0.32\linewidth]{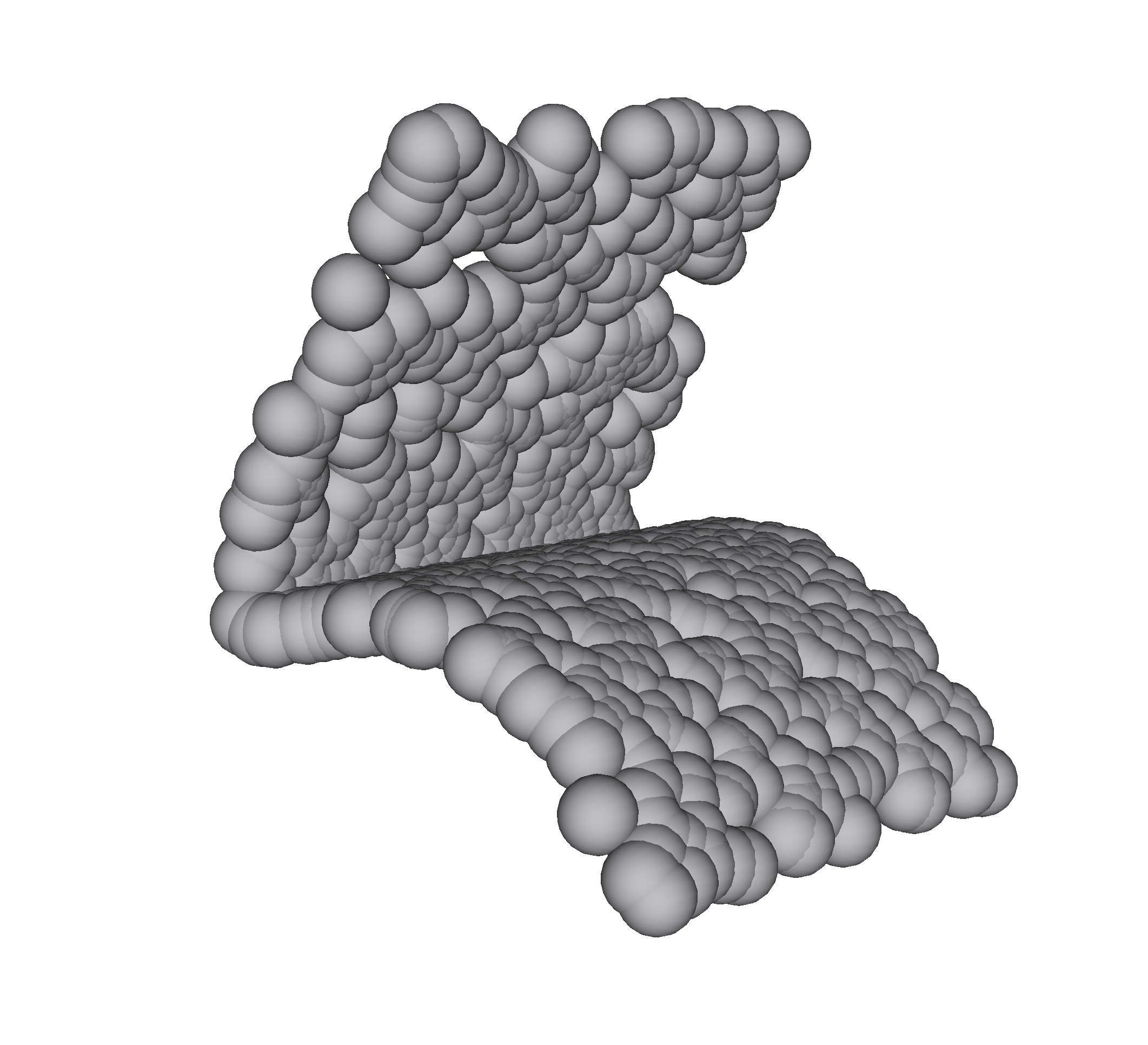}
    \includegraphics[width=0.32\linewidth]{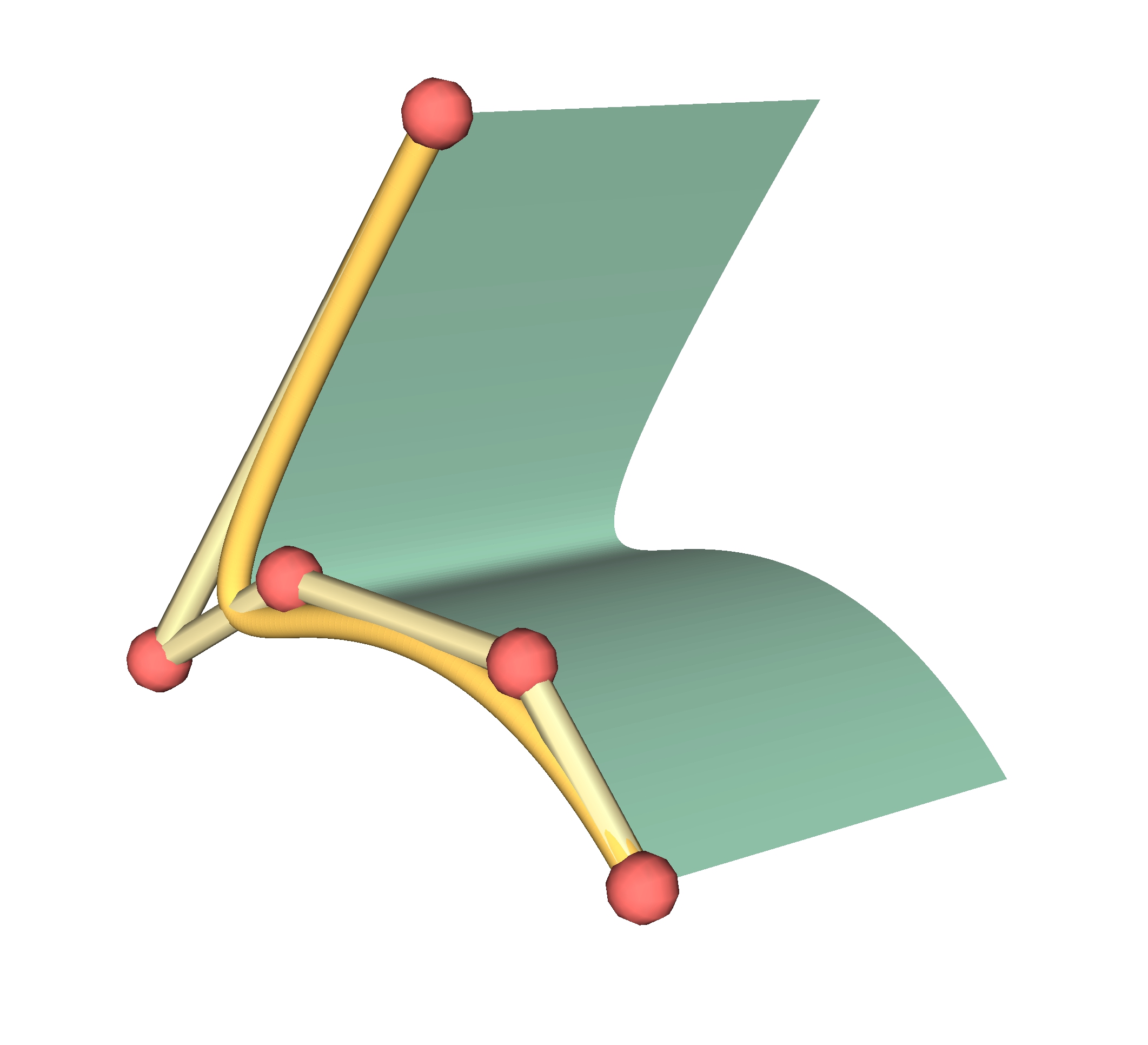}
    \includegraphics[width=0.32\linewidth]{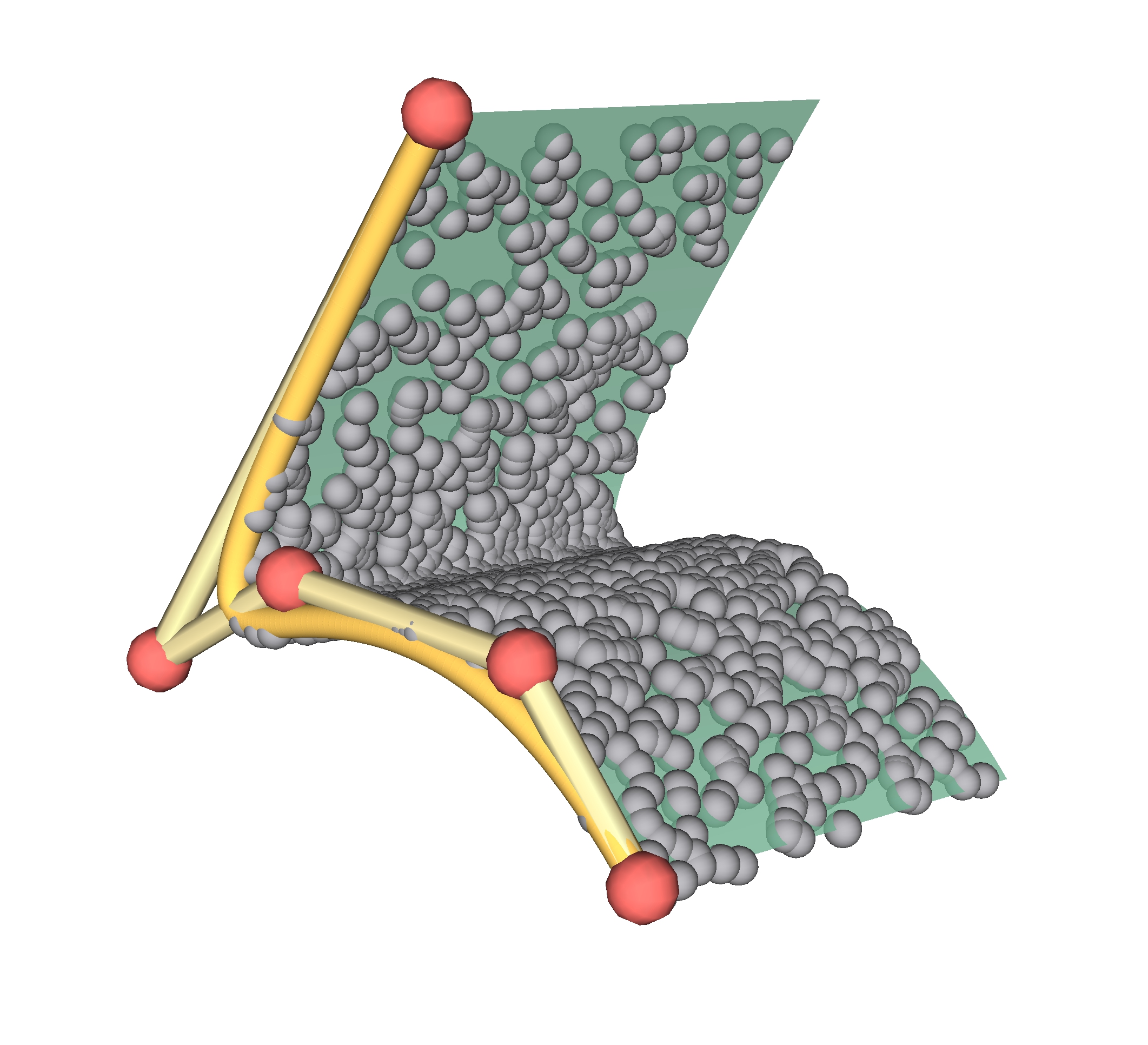}
  \end{minipage}
  \hfill \mbox{}
  \caption{Randomly picked results of point cloud reconstruction. The left column consists of input point clouds, the middle column consists of generator splines revolved (top) or extruded along a path (bottom) to generate surface reconstructions. We plot the input point clouds and predicted surfaces in the right column.}
  \label{results_point_cloud}
\end{figure}

\begin{figure*}[tbp]
  \includegraphics[width=\textwidth]{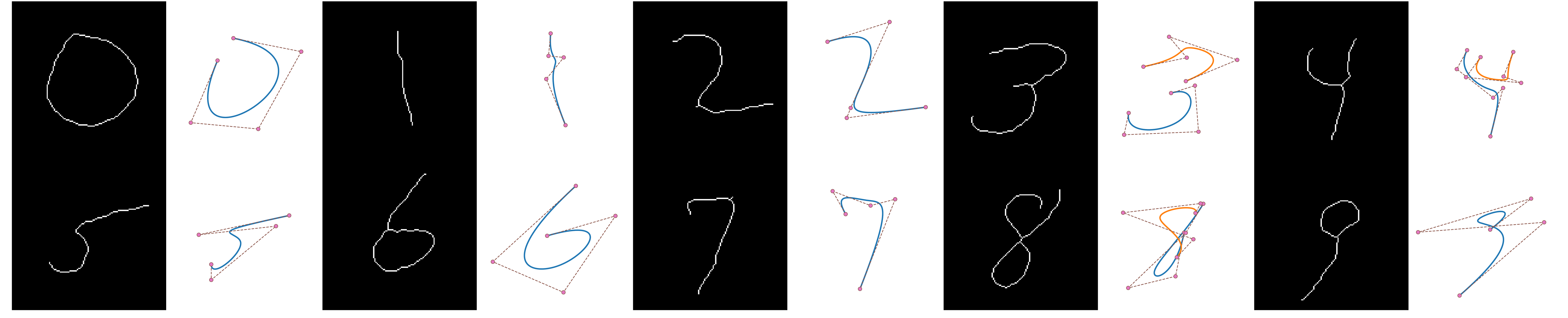}
  \hfill \mbox{}
  \caption{MNIST reconstruction. Reconstruction of MNIST numerals $0$-$9$ is shown. The left sub-image of each of the images is the original MNIST numeral, and the right sub-image is the reconstructed image from our method. The mode that is used is the one with variable number of curves and fixed number of control points for each curve.}
  \label{results_mnist}
\end{figure*}

The optimization results after the RNN-based initialization are an order of magnitude better than the optimization results after a random initialization. This can be quantitatively validated over the entire test set, that has been synthetically generated with random positions of control points and variability in the number of spline curves and control points. This evaluation is indicated in Table~\ref{quant_results_chamfer} as well, where V refers to the case of using variable number of control points on single spline curve, M refers to the case of using multiple spline curves with fixed number of control points and MV refers to the case of using multiple spline curves each containing a variable number of control points.

We also show other quantitative measures in Table ~\ref{quant_results}. Here, we compute the Mean-Squared Error in curve prediction and accuracy in computing number of points and number of curves, as well as the Chamfer distance between the reconstructed prediction and the target curves. It is to be noted that in the case of variable curves with fixed control points, the point accuracy is not an applicable measure since it is known beforehand, and the same holds for curve accuracy in the case of variable control points for single curve. It is observed that in spite of the fact that the performance of curve accuracy and point accuracy both drop in the case of multiple spline curves with variable control points, the drop is very minor and the performance is still excellent by all comparable measures as shown in Table~\ref{quant_results_chamfer}.
\begin{table}[!hbp]
\centering
\begin{tabular}{c|ccc}
\hline
\hline
& NN & NN Init & Random Init \\
\hline
V &  $1.38 \times 10^{-3}$ & $2.30 \times 10^{-5}$ & $3.33 \times 10^{-4}$\\
M &  $1.93 \times 10^{-3}$ & $7.74 \times 10^{-5}$ & $8.25 \times 10^{-4}$\\
MV & $2.22 \times 10^{-3}$ & $9.53 \times 10^{-5}$ & $7.39 \times 10^{-4}$\\
\hline
\end{tabular}
\caption{Chamfer distance between the predicted curve and target curve. ``NN'' denotes the average distance of target from Neural Network prediction, ``NN Init'' denotes the average distance of target from the optimization performed with the Neural Network prediction as initialization, and ``Random Init'' denotes the same distance with optimization performed with random initialization. The optimization here is performed using traditional methods as discussed in Section~\ref{results_sec}. The distances are all normalized to lie in the unit interval. The rows refer to the three modes of variability on which we operate our networks as described in Section~\ref{results_sec}.}
\label{quant_results_chamfer}
\end{table}

\begin{figure}[!htb]
  \centering
  \mbox{} \hfill
  \begin{minipage}{\linewidth}
    \includegraphics[width=0.23\linewidth]{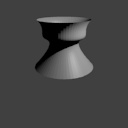}
    \includegraphics[width=0.23\linewidth]{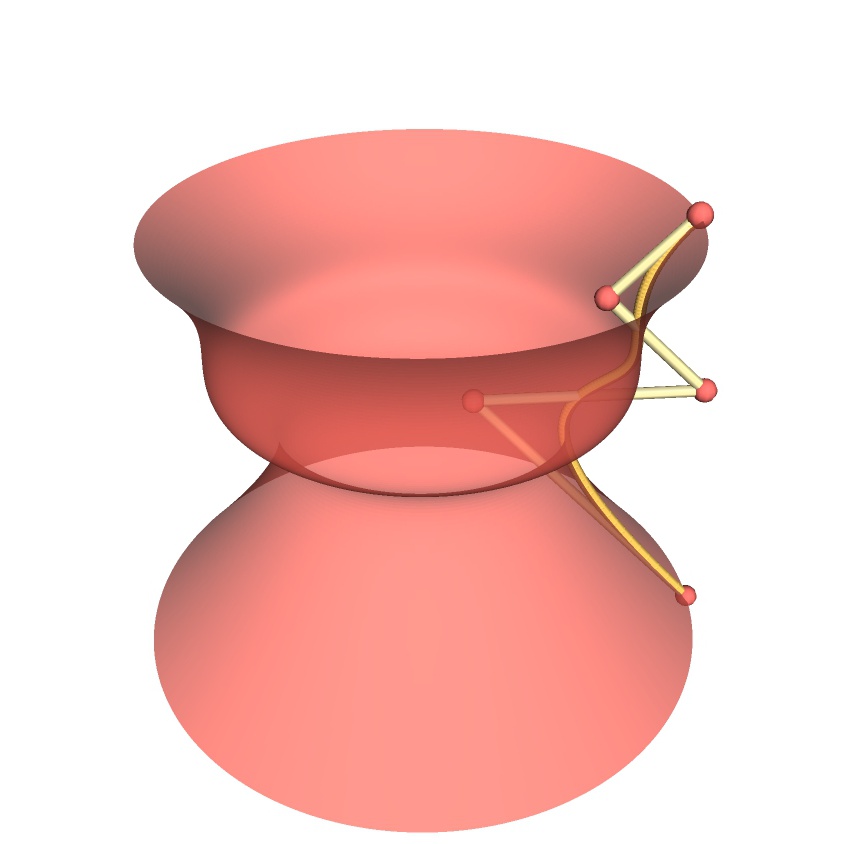}
    \includegraphics[width=0.23\linewidth]{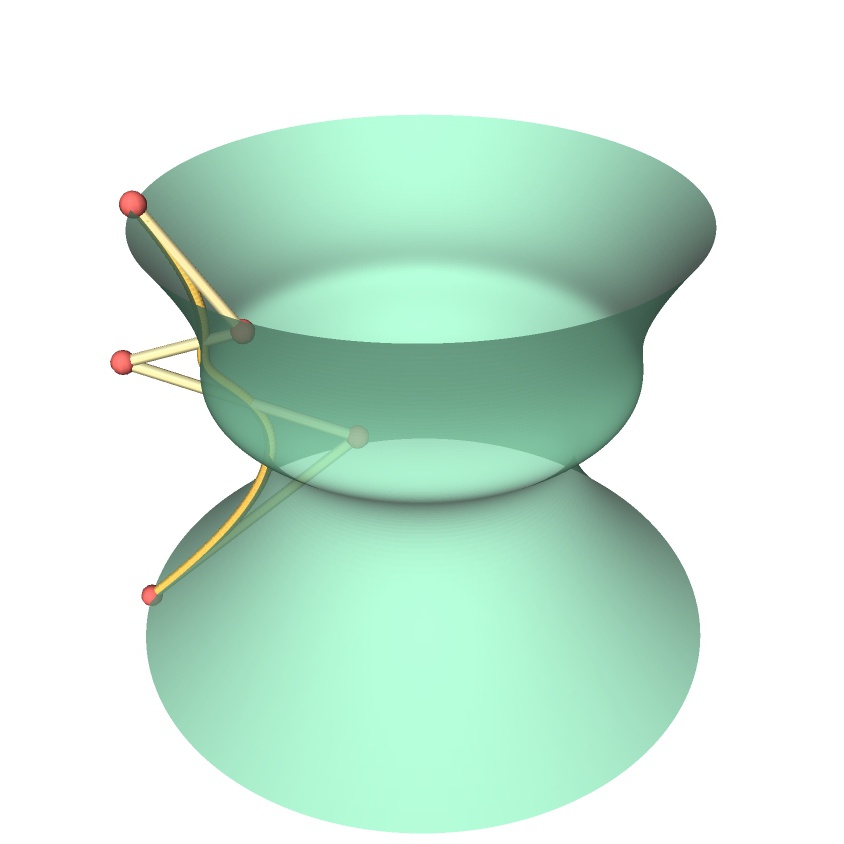}
    \includegraphics[width=0.23\linewidth]{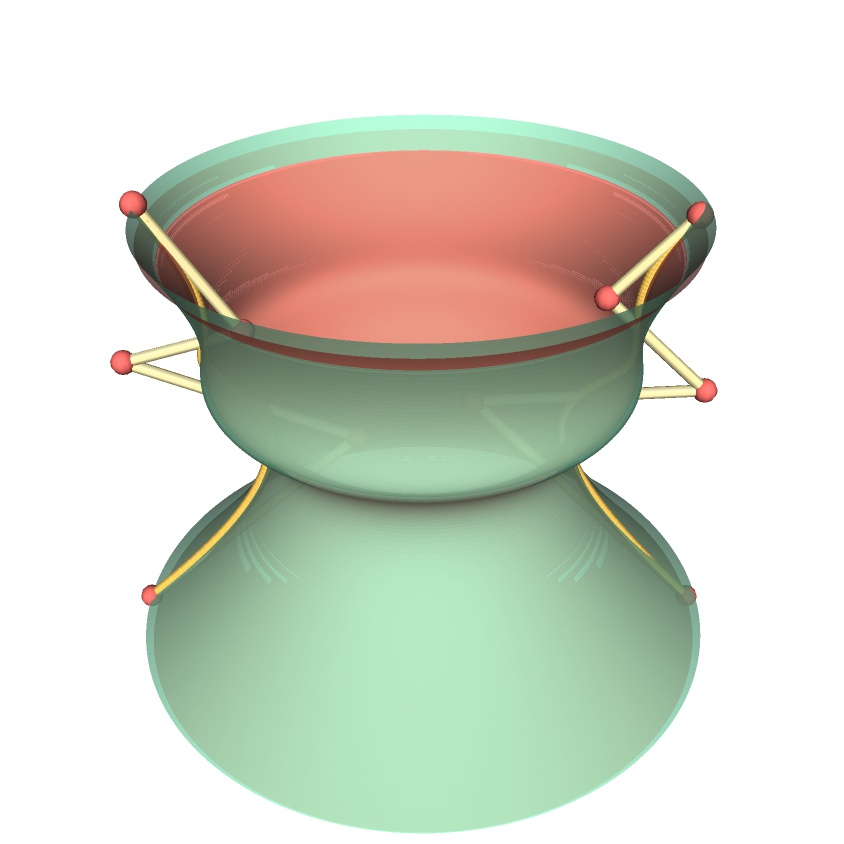}
  \end{minipage}
  \hfill \mbox{}
  \caption{Results of surfaces of revolution on images. Left to Right: Input synthetic image, ground truth surface, predicted surface, ground truth and predicted surface plotted together.}
  \label{results_revolve_surface_img_real}
\end{figure}

\begin{figure}[!htb]
  \centering
  \mbox{} \hfill
  \begin{minipage}{\linewidth}
    \includegraphics[width=0.5\linewidth]{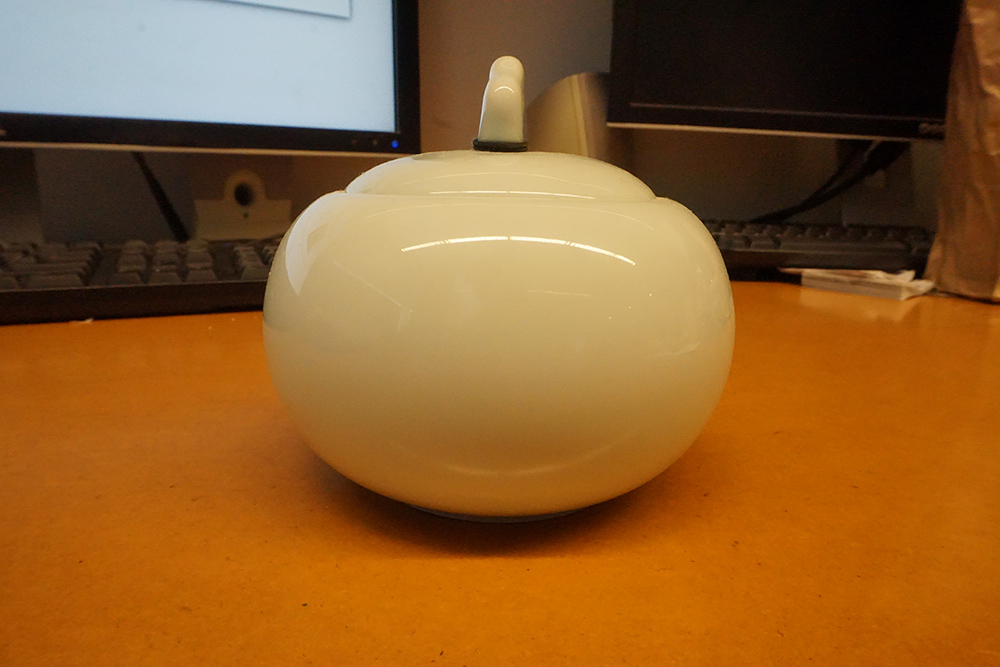}
    \includegraphics[width=0.5\linewidth]{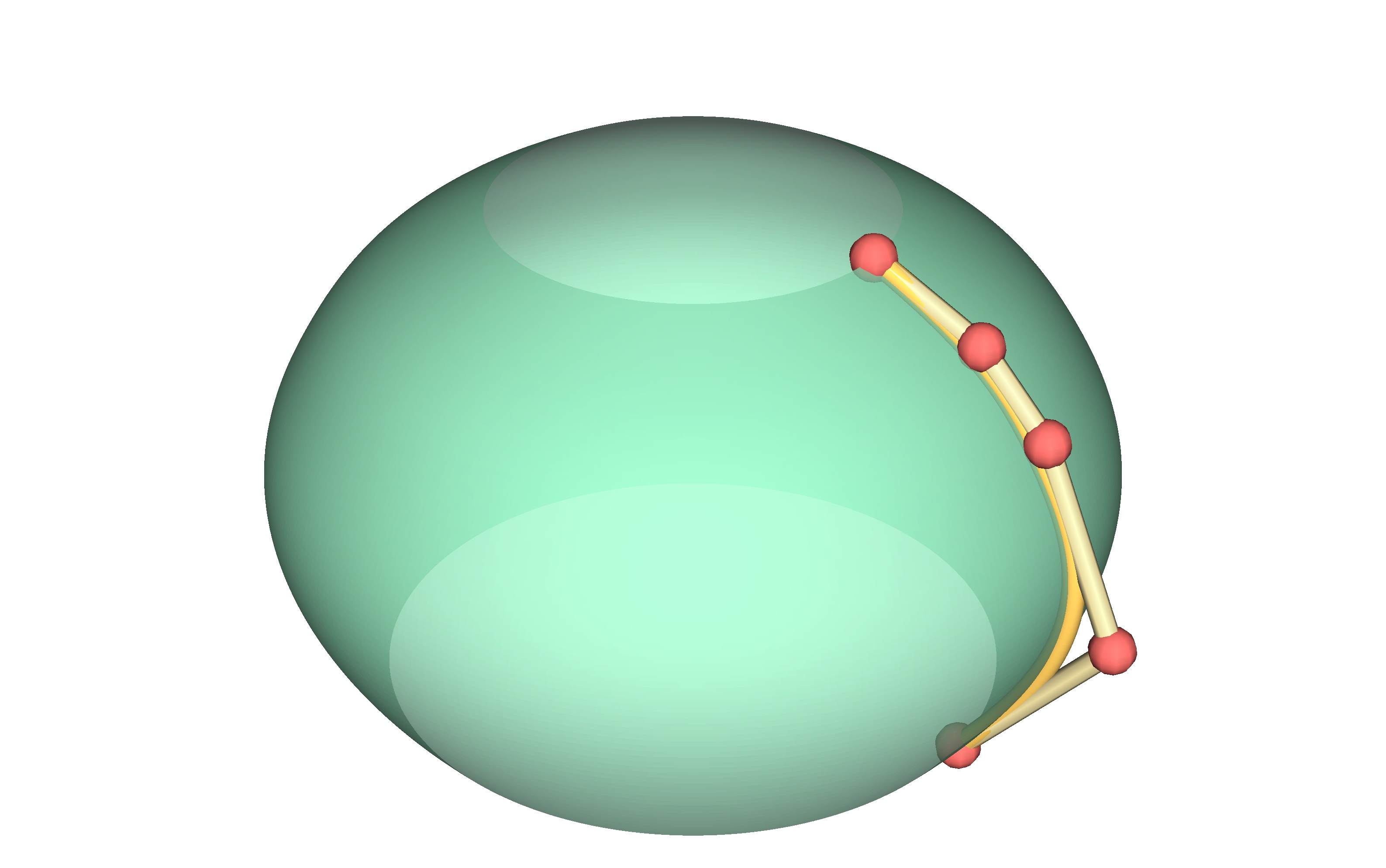}
  \end{minipage}
  \hfill \mbox{}
  \caption{Results of surfaces of revolution on real image. The image on the left is the real image, and one on the right is the reconstructed surface..}
  \label{results_revolve_surface_img_real_2}
\end{figure}

\begin{table}[!hbp]
\begin{tabular}{c|cccc}
\hline
\hline
& MSE & Point Acc & Curve Acc & Chamfer Distance \\
\hline
V & 0.01302 & 94.58 &N/A  & $1.38 \times 10^{-3}$\\
M & 0.02699 & N/A & 99.85 & $1.93 \times 10^{-3}$\\
MV & 0.03738 & 82.75 & 99.49 & $2.22 \times 10^{-3}$\\
\hline
\end{tabular}
\caption{Error and Accuracy measures. Quantitative evaluation of Mean Squared Error, point prediction accuracy, curve prediction accuracy and the Chamfer distance between the predicted and target curves are provided. The accuracy terms are provided in percentages.  The distances are all normalized to lie in the unit interval. The rows refer to the three modes of variability on which we operate our networks as described in Section~\ref{results_sec}.}
\label{quant_results}
\end{table}

\subsubsection{Point Cloud Reconstruction}

The reconstruction of surfaces of extrusion or revolution from inputs in the form of point clouds is performed as described in Section~\ref{real_recons_sec}. A sampling of these reconstructions is shown in Figure~\ref{results_point_cloud}, where spline curve prediction to resemble the point cloud when revolved or extruded is performed in an unsupervised manner.

\begin{figure}[tbp]
  \centering
  \mbox{} \hfill
  \includegraphics[width=\linewidth]{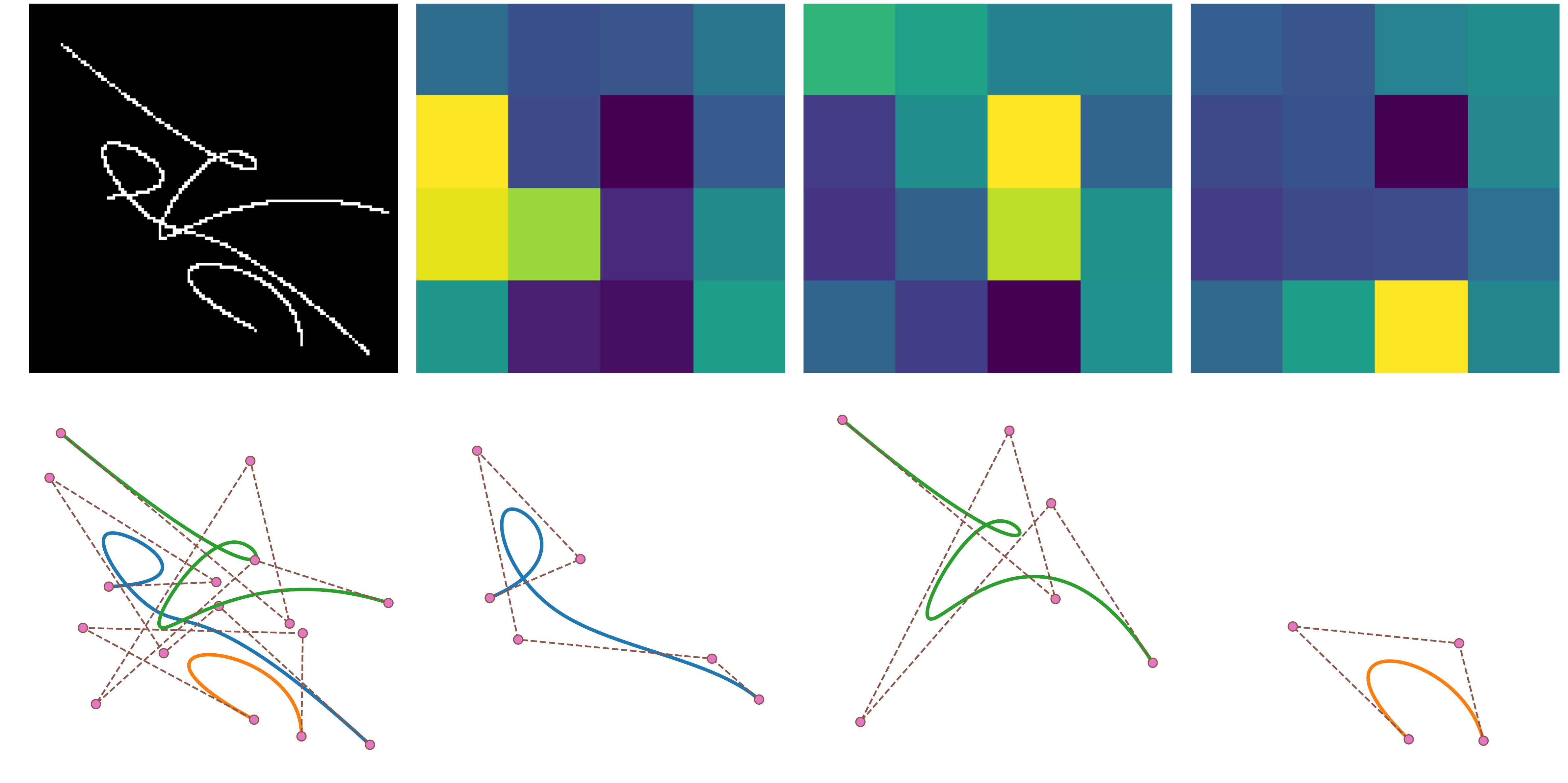}
\vspace{-15pt}
  \hfill \mbox{}
  \caption{Attention results for Hierarchical RNN. The first row shows the input image, and the maps provided by the attention networks for the prediction of each of the individual spline curves. The second row shows the target curves and the predictions of spline curves one at a time, corresponding to the attention map provided in the first row.}
  \label{attention_results}
\end{figure}

\subsubsection{Real/Synthetic Image Reconstruction}

\textbf{MNIST Spline Reconstruction.}
We use the trained networks to reconstruct images in the MNIST dataset \cite{lecun1998gradient}. The networks have been trained on synthetically generated data, and testing it to perform real data reconstruction, as is the case of the MNIST dataset, is overreaching of the capabilities of the network. Nevertheless, it is to be noted that we perform significantly well on this dataset, especially when the input images are more curved in nature, with less sharp edges. A sample of these reconstructed results are shown in Figure~\ref{results_mnist}. 

We first enlarge the MNIST image to size $128 \times 128$, and then thin the digits to lines, which are the input to our network. It is observed that the mode of multiple spline curves with fixed number of control points performs best for this task. It is also to be noted that no post-processing is performed on the result.

\begin{figure}[!htb]
  \centering
  \mbox{} \hfill
  \includegraphics[width=0.95\linewidth]{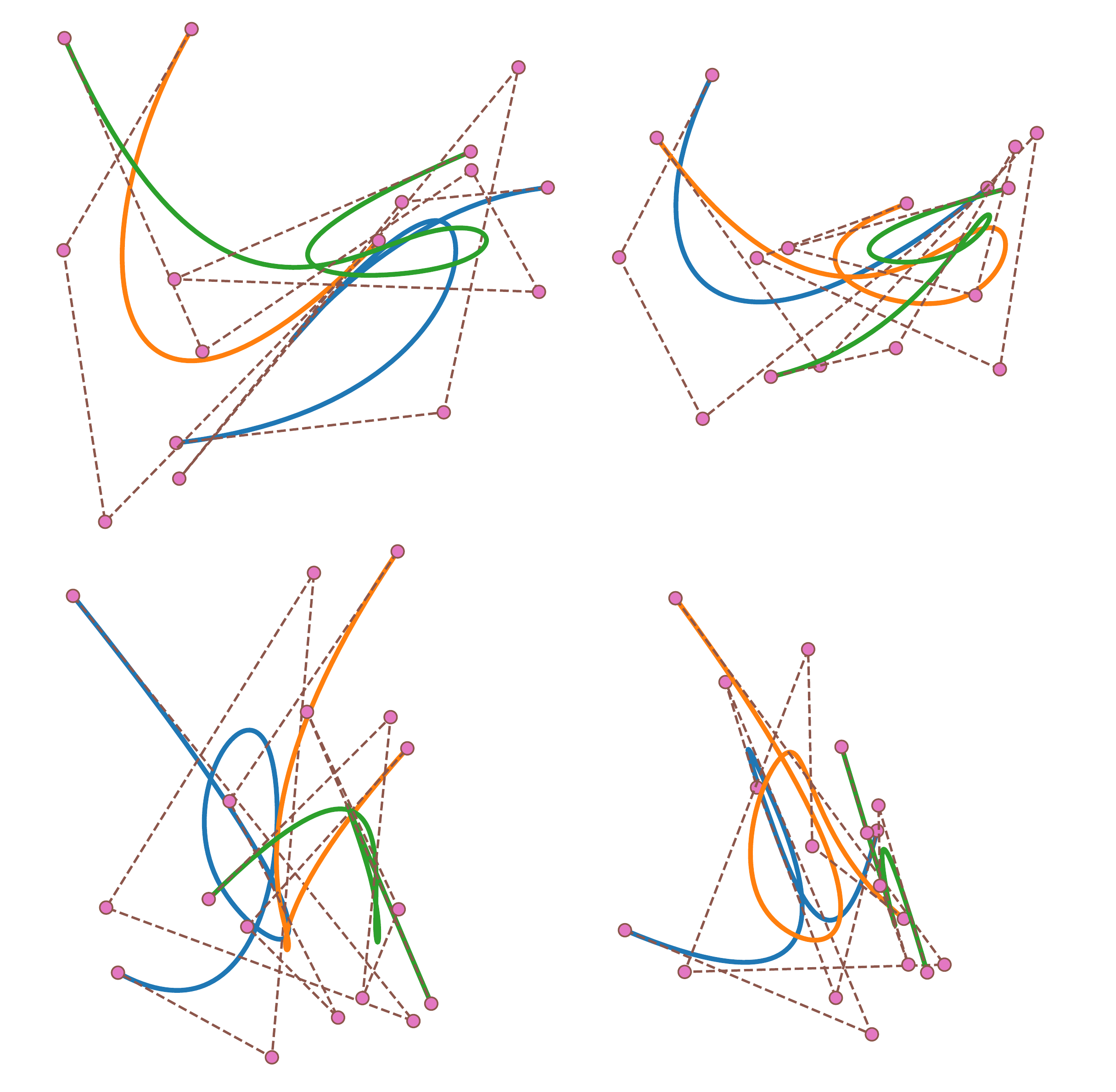}
  \hfill \mbox{}
  \caption{Reconstruction failure. Failure cases of multiple spline curves with variable number of control points}
  \label{results_multiple_variable_failure}
\end{figure}

\textbf{Surface Reconstruction.}
We perform testing on real images to generate surfaces of revolution. As a preprocessing step, we convert the photo to a gray-scale image, crop and pad it properly to fit the input size of the neural network. An instance of this reconstruction, performed as described in Section~\ref{real_recons_sec} on a synthetic input image, is shown in Figure~\ref{results_revolve_surface_img_real}. This can also be performed on real images as illustrated in Figure~\ref{results_revolve_surface_img_real_2}.

\subsubsection{Visualization and Analysis}
The attention map in the Hierarchical RNN is plotted at different curve-prediction iterations in Figure~\ref{attention_results}. This gives us a method to visualize how the networks learns to pick the next curve to predict. In each iteration, the network tends to focus on the regions of the image that correspond to the curve that is being predicted currently. The attention region in these maps is usually close to the center of the curve. Since the operation of convolution is repeated applied, as the network gets deeper, the center part of an object or curve tends to contain more information than the marginal parts, and thus this part deserves more attention.

\subsubsection{Failure Cases}
We showcase a number of failure cases for our method in Figure~\ref{results_multiple_variable_failure}. It is observed that when multiple spline curves are heavily entangled, the model seems to fail. These entanglements are difficult to separate through manual human supervision, and so it can be expected with fairly high chance that this would fail, and this is what is observed in the figure.

\begin{figure}[tbp]
  \includegraphics[width=\linewidth]{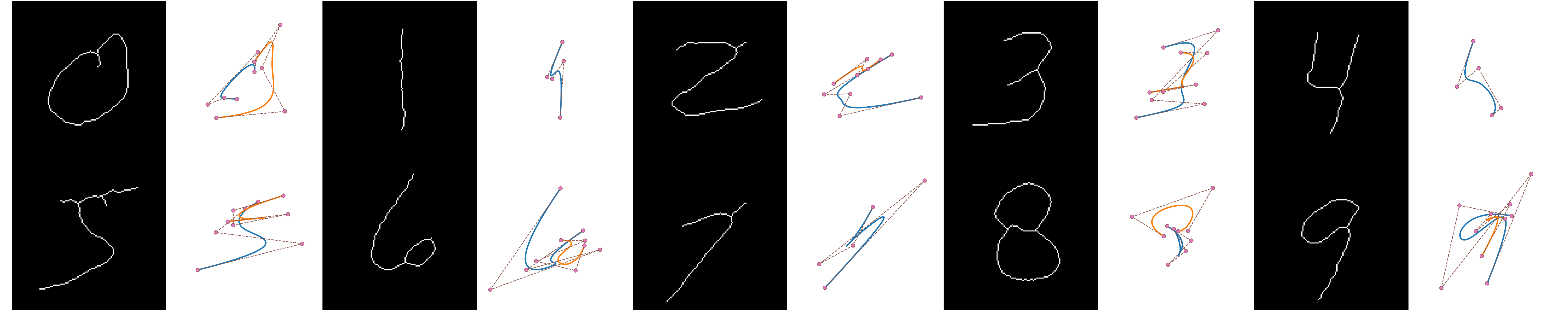}
\vspace{-15pt}
  \hfill \mbox{}
  \caption{Failure cases at MNIST reconstruction. Reconstruction of MNIST numerals $0$-$9$ is shown. The left sub-image of each of the images is the original MNIST numeral, and the right sub-image is the reconstructed image from our method. The mode that is used is the one with multilple number of curves and variable number of control points for each curve.}
  \label{results_mnist_failure}
\end{figure}

We also showcase some failure cases in the reconstruction of MNIST numerals in Figure~\ref{results_mnist_failure}. This can be attributed to the lack of training on real images. There are multiple issues with using the MNIST images as test images. One issue that we run into often is that of closed curves. The training data does not contain closed spline curves, but in MNIST closed curves are ubiquitous, such as numerals 0,6,8,9 in the Figure~\ref{results_mnist_failure}. This tends to be alien to the network when it is seen in the MNIST images. Another issue is that images such as numerals, which are sharper in nature, than plain spline curves, and thus the curves that are predicted, while they attempt to approximate the input image, they are not expressive enough to be able to make a close enough approximation. Real images also contain some noise, such as unnecessary pixels, which might lead the network to predict complicated curves to account for the noise (numerals 1, 2, 5, 7 in the Figure~\ref{results_mnist_failure}). But we believe these issues could be mitigated with some domain adaptation techniques.

There are a few limitations to our methods. For 2D reconstruction, our methods only consider the family of spline curves where the knot position in the splines is fixed. If both the control points and the knots need to vary, our methods need modifications. For 3D reconstruction, our methods only consider surfaces with no self-intersections and decreasing y-coordinates to avoid the local minima problem. Finally, due to lack of training data, we train our model on synthetic data and test it on real data. While we do this, we do not apply domain adaptation techniques to help the network generalize well to real data, but this is a problem which could be attempted as a future work.

\section{Conclusion and Future Work}
In this paper, we have illustrated approaches to reconstruct spline curves and surfaces using data driven approaches.
Being both different and complementary to the traditional methods of spline fitting, our methods, adaptive to different forms of inputs, do not need initialization and can handle variable number of control points.
There are many exciting directions that remain to be explored.
A viable future direction would be to investigate methods to detect, decompose, and recover mutiple parametric surfaces from single images and consistently assemble them across multiple images.
Another area of interest would be to study how other types of information, such as semantics and physics, can be utilized to design and reconstruct clean, complex, and functional geometry and structures.

\bibliographystyle{eg-alpha}

\bibliography{parametric}
\end{document}